\documentclass{article}

\pdfoutput=1
\usepackage{Others/PRIMEarxiv}

\usepackage[utf8]{inputenc} 
\usepackage[T1]{fontenc}    
\usepackage{url}            
\usepackage{booktabs}       
\usepackage{amsfonts}       
\usepackage{amsmath}
\usepackage{amssymb}
\usepackage{mathtools}

\DeclareMathOperator*{\argmin}{arg\,min}
\usepackage{algorithm}
\usepackage{algpseudocode}

\usepackage{lscape}
\usepackage{nicefrac}       
\usepackage{microtype}      
\usepackage{lipsum}
\usepackage{fancyhdr}       
\usepackage{graphicx}       
\usepackage{xcolor}
\usepackage{setspace}
\usepackage{chngcntr}

\usepackage{biblatex}
\usepackage[pdftex]{hyperref}

\title{ Posterior sampling with CNN-based, Plug-and-Play regularization with applications to Post-Stack Seismic Inversion}

\author{
  Muhammad Izzatullah\thanks{\textbf{Corresponding author:} \texttt{muhammad.izzatullah@kaust.edu.sa}}, Tariq Alkhalifah, Juan Romero, Miguel Corrales, Nick Luiken, Matteo Ravasi \\
  King Abdulllah University of Science and Technology (KAUST) \\}

\begin{document}
\maketitle
\doublespacing

\begin{abstract}
Uncertainty quantification is crucial to inverse problems, as it could provide decision-makers with valuable information about the inversion results. For example, seismic inversion is a notoriously ill-posed inverse problem due to the band-limited and noisy nature of seismic data. It is therefore of paramount importance to quantify the uncertainties associated to the inversion process to ease the subsequent interpretation and decision making processes. Within this framework of reference, sampling from a target posterior provides a fundamental approach to quantifying the uncertainty in seismic inversion. However, selecting appropriate prior information in a probabilistic inversion is crucial, yet non-trivial, as it influences the ability of a sampling-based inference in providing geological realism in the posterior samples. To overcome such limitations, we present a regularized variational inference framework that performs posterior inference by implicitly regularizing the Kullback-Leibler divergence loss with a CNN-based denoiser by means of the Plug-and-Play methods. We call this new algorithm Plug-and-Play Stein Variational Gradient Descent (PnP-SVGD) and  demonstrate its ability in producing high-resolution, trustworthy samples representative of the subsurface structures, which we argue could be used for post-inference tasks such as reservoir modelling and history matching. To validate the proposed method, numerical tests are performed on both synthetic and field post-stack seismic data.
\end{abstract}
\vspace{0.5cm}
\section{Introduction}
Seismic inversion addresses the problem of estimating unknown subsurface model parameters from acquired seismic data. Such an inverse problem is ill-posed due to presence of noise in the recorded seismic data, the use of inaccurate physics, and the inherent nullspace in the forward modeling operator, which leads to a non-uniqueness in the expected solution ~\cite{menke2018seismic}. Typically, seismic inversion is cast as an iterative, optimization problem that attempts to minimize the difference between observed and modeled seismic data. Alternatively, this optimization problem can be setup in a probabilistic fashion by leveraging the Bayesian formulation to find a distribution of solutions consistent with the observed seismic data -- known as the posterior distribution. In such a way, one can obtain a more comprehensive description of the inverted model parameters. In particular, practitioners can further quantify the solution's uncertainty (i.e., assessing the variability among the possible solutions) by sampling from the posterior distribution and/or computing statistics (e.g., mean, standard deviation, marginal probabilities)~\cite{tarantola2005seismic}.

Over the past decade, probabilistic inference has gained momentum in the geophysics community as a tool to quantify uncertainties in seismic inversion. However, such a procedure often requires high-dimensional posterior distribution sampling, which is prohibitively computationally expensive. As such, the fully probabilistic inference based on the Bayesian formulation is often reduced, for example, to the Laplace approximation approach~\cite[e.g.,][]{bui2013laplace, zhu2016laplace, liu2019laplace, izzatullah2019laplace, izzatullah2021laplace, izzatullah2022laplace} to alleviate such limitations. The Laplace approximation is the simplest family of posterior approximations for high-dimensional problems (e.g., seismic inversion), as it forms a Gaussian approximation to the exact posterior in the locality of the maximum a posteriori (MAP) solution. In this case, the covariance equals the negative inverse Hessian (i.e. approximation thereof) of the loss functions evaluated at the MAP solution. The quality of the Laplace approximation depends on its covariance estimation as it quantifies the uncertainty representation, and there are various approaches to approximate the covariance~\cite[e.g.,][]{spantini2015laplace, ritter2018laplace, daxberger2021laplace}. However, a good covariance approximation is computationally limited by the number of the solution's parameters. In principle, the Laplace approximation approach is also known as a poor man's Bayesian inference~\cite{tzikas2008laplace}, i.e., it is a way of cheaply including prior knowledge by compensating for the high price of evaluating the full posterior probability density (e.g., for high dimensional problems). However, it often only quantifies uncertainty accurately in the proximity of the MAP solution.

Recently, the potential of gradient-based Markov Chain Monte Carlo (MCMC) methods for probabilistic seismic inversion is also being explored in the geophysics community. For example, the Langevin Monte Carlo (LMC)~\cite{welling2011lmc, izzatullah2020stein, siahkoohi2020lmc, izzatullah2021lmc} and Hamiltonian Monte Carlo (HMC)~\cite[e.g.,][]{sen2016hmc, fichtner2019hmc, gebraad2020hmc, fichtner2021hmc, zunino2022hmc} are used in probabilistic seismic inversion to sample from high-dimensional posterior distributions. LMC and HMC sit at the boundary between gradient-based optimization and derivative-free MCMC methods (e.g., random walk), thus making them robust in sampling high-dimensional distributions from the gradient information. Despite their robustness, MCMC sampling methods require a large number of sampling steps and a long burn-in period to perform an accurate probabilistic inference due to their sequential nature (i.e., Markov chain property), which hinders their applicability to high-dimensional large-scale problems (e.g., seismic inversion) due to the costs associated with the forward and adjoint operators evaluation per iteration.

Lately, variational inference (VI) approaches have gained increasing interest in the scientific community as an alternative approach to tackle the challenges of probabilistic inference in high-dimensional inference problems, such as seismic inversion. VI approximates the posterior distribution through a surrogate distribution that is easy to sample (e.g., Gaussian distribution). This approach frames the computationally costly sampling procedure of MCMC into an optimization problem in the space of probability distributions, i.e., a functional minimization problem of Kullback-Leibler (KL) divergence between the surrogate and posterior distributions. This surrogate distribution is then used for conducting Bayesian inference. We refer the reader to~\cite[ch.~8]{santambrogio2015ot} for an introduction to optimization in the probability space. VI can be solved through several approaches, e.g., variational Bayes~\cite[e.g.,][]{blei2017vi, nawaz2020viseismic, urozayev2022viseismic}, deep neural network representation~\cite[e.g.,][]{baptista2020vi, kruse2021vi, siahkoohi2021vi, siahkoohi2022vi, zhao2022viseismic} or particle evolution update such as Stein Variational Gradient Descent (SVGD)~\cite[e.g.,][]{zhang2020svgdseismic, zhang2021svgdseismic, smith2022svgd, corrales2022svgd}. Variational inference provides a balance between the Laplace approximation and MCMC methods because it provides more flexibility in approximating the posterior distributions compared to the former at a reduced computational cost compared to MCMC methods, especially in high-dimensional large-scale inverse problems.

In this work, we focus on advancing the frontier of variational inference through the SVGD algorithm. We develop the \emph{regularized variational inference} framework where the minimization of KL divergence loss between the surrogate and the target posterior distribution is regularized by a general regularization function. This framework provides flexibility in incorporating prior information on the target distribution compared to the traditional approach through the Bayesian formulation. More specifically, we propose the \emph{Plug-and-Play Stein Variational Gradient Descent (PnP-SVGD)} -– a novel VI algorithm for sampling from a regularized target posterior distribution, where the regularizer is a convolutional neural network (CNN) based denoiser.

\subsection{Related works}
The seminal work of Venkatakrishnan et al.~\cite{venkatakrishnan2013pnp} introduced the Plug-and-Play (PnP) framework in 2013. This framework provides flexibility in regularizing inverse problems with priors that do not necessarily possess an easy-to-compute proximal mapping. More precisely, ~\cite{venkatakrishnan2013pnp} interpreted a proximal mapping as a denoising inverse problem, suggesting to replace it with any state-of-the-art denoising method whilst solving a deterministic inverse problem with variable splitting algorithms (e.g., alternating direction method of multipliers -- ADMM). Whilst ~\cite{venkatakrishnan2013pnp} initially suggested to use statistical denoisers such as BM3D~\cite{dabov2007bm3d} or sparse learned dictionaries such as K-SVD~\cite{aharon2006ksvd}, this framework has also opened ways to utilize novel deep-learning based denoising algorithms. For example, Zhang et al.~\cite{zhang2017pnp, zhang2021pnp} extend the PnP framework using a learned prior where the denoiser is discriminatively learned via a deep CNN with large modeling capacity. Such extensions proved that the deep denoiser prior not only significantly outperforms other state-of-the-art model-based methods but also achieves competitive or superior performance against state-of-the-art learning-based methods in image restoration tasks. In 2019, Ryu et al.~\cite{ryu2019pnp} analyzed the convergence of PnP framework algorithms under a Lipschitz assumption on the deep learning-based denoiser and proposed spectral normalization in training deep learning-based denoisers to enforce the proposed Lipschitz condition for a guaranteed convergence. The PnP framework is further developed for Bayesian inference by incorporating it into the LMC through Tweedie's identity~\cite{laumont2022pnp}. The PnP framework has demonstrated its potential in geophysical applications by solving seismic deblending~\cite[e.g.,][]{bahia2022seismicpnp, luiken2022seismicpnp}, data interpolation~\cite[e.g.,][]{zhang2020seismicpnp, xu2022seismicpnp}, and seismic inversion ~\cite{romero2022pnp}. More specifically, Romero et al.~\cite{romero2022pnp} present a comparison between standard model-based and deep-learning-based regularization techniques in post-stack seismic inversion and provide insights into the optimization process as well as denoiser-related parameters tuning. Their result suggests that implicit priors represented by pre-trained denoising neural networks can effectively drive the solution of seismic inverse problems towards high-resolution models.

In this work, we focus on developing a framework for probabilistic geophysical inversion applications, which we later apply to a post-stack seismic inversion problem to quantify its model uncertainty. Our work extends the framework of the posterior regularization introduced by Altun and Smola~\cite{altun2006reg} and Zhu et al.~\cite{zhu2014reg}, which performs posterior inference with regularization terms. In principle, our work shares similarities with the works of Zhang et al.~\cite{zhang2022svgd} and Liu et al.~\cite{liu2021svgd}, where both propose a family of constrained sampling algorithms that generalize LMC and SVGD to incorporate a constraint specified by a general nonlinear function. However, we develop a posterior regularization framework that can accomodate a CNN-based denoiser regularization using insights from the PnP deterministic framework; as a result, our method provides improved probabilistic inversion solutions in geophysical applications compared to methods that use generic handcrafted regularization techniques. With this insight, we develop a novel sampling algorithm based on SVGD for sampling from a regularized target posterior distribution. Other related works are proposed by Shi et al.~\cite{shi2021svgd}, where they developed an SVGD algorithm for constrained domains and non-Euclidean geometries using the mirror-descent map, and by Li et al.~\cite{li2022svgdmoll} where they introduced a new optimization-based method for sampling called mollified interaction energy descent (MIED), as an alternative to SVGD. From the theoretical development side, He et al.~\cite{he2022svgdreg} introduced regularized SVGD, which interpolates between the Stein Variational Gradient Flow and the Wasserstein Gradient Flow.

\subsection{Main contributions}
The contributions of this work to the field under study are as follows:
\begin{enumerate}
    \item We propose PnP-SVGD -- a novel probabilistic algorithm for sampling from a regularized target posterior distribution.
    \item Inspired by the contributions in \cite{altun2006reg, zhu2014reg}, we present a CNN-based PnP regularized variational inference framework for post-stack seismic inversion. This computational framework performs posterior inference by regularizing the KL divergence loss with a CNN-based denoiser PnP regularization \cite{venkatakrishnan2013pnp, zhang2017pnp, ryu2019pnp, zhang2021pnp, romero2022pnp}.
    \item We demonstrate the practicality and robustness of our proposed methodology with  numerical examples of post-stack seismic inversion based on both a realistic synthetic model and the Volve field data.
\end{enumerate}

\subsection{Outline}
The outline of the paper is as follows: First, we introduce the theoretical framework by formulating the post-stack seismic inversion within a Bayesian framework. We then provide an overview of the deterministic PnP formulation. This is followed by a description of the general variational inference framework and its regularized counterpart. Next, we provide an overview of the SVGD algorithm and introduce our novel PnP-SVGD algorithm that allows sampling from a regularized target posterior distribution. We demonstrate the effectiveness of the proposed methodology through numerical examples, including using a realistic synthetic dataset and the Volve oilfield dataset.
\section{Theoretical Framework}
Our objective is to present a framework that enables the inclusion of deep learning-based regularization into the variational inference methodology. This enables robust inference with high-resolution posterior samples, which are geologically plausible representations of the subsurface structures of interest. We note that most of the common probabilistic inference procedures currently lack such a feature because they trade uncertainty quantification capabilities with the ability of including highly informative priors.
%
\subsection{Bayesian formulation} \label{sec:theory-bayes}
For linear problem the relationship between a model and the observed data can be described as follows:
\begin{equation} \label{eq:post-stack-data}
    \mathbf{d} = \mathbf{G}{\mathbf{m}} + \boldsymbol\varepsilon
\end{equation}
where $\mathbf{G}$ is the modeling operator. For our post-stack seismic inversion problem $\mathbf{G} = \mathbf{W} \mathbf{D}$, where $\mathbf{D}$ is a first-order derivative operator, and $\mathbf{W}$ is a convolutional operator with the seismic wavelet $w$. Here, $\mathbf{m}$ is natural logarithm of the AI model associated with the observed data $\mathbf{d}$ ~\cite{stolt1985}, and $\boldsymbol\varepsilon$ is a vector of measurement noise, which includes errors in the seismic wavelet approximation. Due to the band-limited nature of the wavelet, the operator $\mathbf{G}$ presents a nullspace as it filters out all low and high frequencies outside of the spectrum of the wavelet itself. Therefore, framing the seismic inversion problem into the Bayesian framework allows us to characterize the variability of all plausible solutions that equally match the observed data within the uncertainty of the observations.

The Bayesian framework for the post-stack seismic inversion can be formulated through Bayes' Theorem~\cite[ch.~3.1.2]{ProbMLBook2}, where the posterior ($\mathbf{p}(\mathbf{m}|\mathbf{d})$) can be defined up to its normalizing constant as
\begin{equation}\label{eq:unnormalized-bayes}
    \pi(\mathbf{m}) \triangleq \mathbf{p}(\mathbf{m}|\mathbf{d}) = \frac{1}{Z}\mathbf{p}(\mathbf{d}|\mathbf{m})\mathbf{p}(\mathbf{m}).
\end{equation}
where $Z$ is the normalization constant that is independent of $\mathbf{m}$. The Bayesian framework stated above introduces the notion of prior probability density ($\mathbf{p}(\mathbf{m})$) and the likelihood function  ($\mathbf{p}(\mathbf{d}|\mathbf{m})$). Here, the prior probability density encodes the confidence of the prior information of the unknown AI model $\mathbf{m}$, whereas the likelihood function describes the conditional probability density of the AI model to model the observed seismic data. Based on Bayes' theorem, we obtain the posterior probability density of the AI model given the observed data, $\mathbf{p}(\mathbf{m}|\mathbf{d})$, by combining the prior probability density and the likelihood function. In the optimization problem where the MAP solution is sought, equation~\eqref{eq:unnormalized-bayes} can be reformulated as a minimization of the negative log-posterior distribution as follows:
\begin{equation}\label{eq:log-posterior-minimization}
\begin{split}
    \mathbf{m}^{\ast} &= \argmin_{\mathbf{m}} -\log\pi(\mathbf{m})\\
    &= \argmin_{\mathbf{m}} -\log\mathbf{p}(\mathbf{d}|\mathbf{m}) - \log\mathbf{p}(\mathbf{m}) + \log Z\\
    &= \argmin_{\mathbf{m}} -\underbrace{\log\mathbf{p}(\mathbf{d}|\mathbf{m})}_{\text{log-likelihood}} - \underbrace{\log\mathbf{p}(\mathbf{m})}_{\text{log-prior}},
\end{split}
\end{equation}
where the constant term $\log Z$ in the above equation can be ignored without affecting the inference/optimization procedure.
%
\subsection{Plug and Play framework for deterministic inversion}
Before combining the PnP framework with probabilistic inversion, we provide an overview of the deterministic PnP formulation using the same log-posterior distribution formulation as in equation~\eqref{eq:log-posterior-minimization}. The goal of deterministic inversion is to seek an optimal solution (i.e., MAP solution in the Bayesian context) that maximizes the log-posterior distribution. In this section, we will interchangeably use the terminology from the optimization and Bayesian inference literature to bridge these two concepts and assist the reader in understanding both formulations.
%
\subsubsection{Plug-and-Play Primal-Dual algorithm}
The Primal-Dual algorithm originally introduced by Chambolle and Pock in their seminal work in 2011~\cite{Chambolle2011convex}, is designed to solve non-smooth convex optimization problems of the form:
\begin{equation}\label{eq:pnp-fg}
    \min_{\mathbf{m}} f(\mathbf{m}) + g(\mathbf{K}\mathbf{m}),
\end{equation}
where $f(\mathbf{m})$ and $g(\mathbf{K}\mathbf{m})$ are (possibly non-smooth functionals) that, within our context, represent the negative log-likelihood and log-prior in the last line of equation~\eqref{eq:log-posterior-minimization}, respectively. This problem can be transformed into its primal-dual equivalent problem by introducing $\mathbf{z} = \mathbf{K}\mathbf{m}$ (i.e., the variable splitting step~\cite{boyd2004convex, Parikh2013convex, beck2017convex}) and finding the saddle-point of the Lagrangian $\mathcal{L}(x, y, z)$, which leads to the primal-dual algorithm:
\begin{eqnarray}\label{eq:pnp-pd-alg}
    \mathbf{y}_{t+1} & = & (I + \sigma \partial g^{\star})^{-1}(\mathbf{y}_{t} + \sigma \mathbf{K}\hat{\mathbf{m}}_{t})\nonumber\\ 
    \mathbf{m}_{t+1} & = & (I + \tau \partial f)^{-1}(\mathbf{m}_{t} - \tau \mathbf{K}^{\star} \mathbf{y}_{t+1})\\ 
    \hat{\mathbf{m}}_{t+1} & = & \mathbf{m}_{t+1} + \theta (\mathbf{m}_{t+1} - \mathbf{m}_{t})\nonumber
\end{eqnarray}
where the scalars $\sigma$ and $\tau$ act as step-lengths on the subgradient of $f$ and $g$, respectively. As shown in \cite{Chambolle2011convex}, convergence is guaranteed in the primal-dual algorithm when $\sigma \tau L^2 < 1$ where  $L = \Vert \mathbf{K} \Vert _2^2$. The operator $(I + \tau \partial f)^{-1}$ is called the resolvent~\cite[ch.~3.2]{Parikh2013convex} and is equivalent to the proximal operator of $f$~\cite[ch.~4]{Parikh2013convex}, defined as:
\begin{equation}\label{eq:pnp-prox}
    \text{prox}_{\tau f}(\mathbf{m}) = (I + \tau \partial f)^{-1}(\mathbf{m}) = \argmin_{\mathbf{y}} \left(\frac{1}{2 \tau}\Vert \mathbf{m}-\mathbf{y}\Vert_2^2 + f(\mathbf{y})\right),
\end{equation}
Similarly, the proximal operator of the convex conjugate function $g^*(x)$ is obtained via the Moreau identity~\cite[ch.~2.5]{Parikh2013convex}:
\begin{equation}\label{eq:pnp-moreau}
    \text{prox}_{\sigma g^{\star}}(\mathbf{m}) =  \mathbf{m} -  \sigma\:\text{prox}_{g/\sigma}(\mathbf{m} / \sigma ).
\end{equation}
Based on the proximal operator definition, the primal-dual algorithm becomes:
\begin{eqnarray}\label{eq:pnp-pd-1}
    \mathbf{y}_{t+1} & = & \mathbf{y}_{t} + \sigma \mathbf{K}\hat{\mathbf{m}}_{t} - \sigma \text{prox}_{g/\sigma}\left(\mathbf{y}_{t}/\sigma +\mathbf{K}\hat{\mathbf{m}}_{t} \right) \nonumber\\
    \mathbf{m}_{t+1} & = & \text{prox}_{\tau f}(\mathbf{m}_{t} - \tau \mathbf{K}^{\star} \mathbf{y}_{t+1})\\ 
    \hat{\mathbf{m}}_{t+1} & = & \mathbf{m}_{t+1} + \theta (\mathbf{m}_{t+1} - \mathbf{m}_{t}).\nonumber
\end{eqnarray}
Implementing proximal operators is advantageous as these operators often have closed-form solutions, making them computationally efficient when dealing with high-dimensional datasets~\cite{Parikh2013convex, beck2017convex}. Additionally, as first pointed out in~\cite{venkatakrishnan2013pnp}, the proximal operator definition as shown in equation~\eqref{eq:pnp-prox} can be interpreted as an denoising inverse problem, where $\mathbf{m}$ is the noisy signal, whereas $\mathbf{y}$ is the clean signal that we attempt to estimate, and $f(\mathbf{y})$ is a regularization term that enables us to enforce our prior regarding the expected solution. In other words, equation~\eqref{eq:pnp-prox} corresponds to the MAP estimator under the assumption of white, Gaussian noise with $\sigma$ being its variance. Based on this definition of the proximal operator identified as a Gaussian denoiser in MAP framework, \cite{venkatakrishnan2013pnp} proposed the PnP framework where a generic denoiser of choice can be used to replace the proximal operator of the dual-update. This approach is appealing because it can leverage state-of-the-art denoisers like those based on the CNNs. Hence, the PnP-PD algorithm can be written as follows:
\begin{eqnarray}\label{eq:pnp-PnP}
    \mathbf{y}_{t+1} & = & \mathbf{y}_{t} + \sigma \hat{\mathbf{m}}_{t} - \mathrm{H}_{1/\sigma}\left(\mathbf{y}_{t}/\sigma + \hat{\mathbf{m}}_{t}\right) \nonumber\\
    \mathbf{m}_{t+1} & = & \text{prox}_{\tau f}(\mathbf{m}_{t} - \tau\mathbf{y}_{t+1}) \\
    \hat{\mathbf{m}}_{t+1} & = & \mathbf{m}_{t+1} + \theta (\mathbf{m}_{t+1} - \mathbf{m}_{t})\nonumber
\end{eqnarray}
where in this specific case $\mathbf{K}=\mathbf{I}$ and $\mathrm{H}_{1/\sigma}$ is defined as a CNN-based denoiser with $1/\sigma \geq 0$ representing the variance of the noise to be removed. The PnP-PD steps are summarized in Algorithm~\ref{alg:pnp-pd}.
\begin{algorithm}
\caption{PnP-Primal-Dual (PnP-PD)}\label{alg:pnp-pd}
\textbf{Input:} A CNN-Based denoiser $\mathrm{H}_{\sigma}$, an initial AI model $\mathbf{m}_{0}$, and initial Langrange multiplier $\mathbf{y}_{0}=\mathbf{0}$.
\textbf{Output:} The MAP solution $\mathbf{m}_{MAP}$ that approximates the AI model.

\begin{algorithmic}
\For{iteration $t$} 

\State {$\mathbf{y}_{t+1} \longleftarrow \mathbf{y}_{t} + \sigma \hat{\mathbf{m}}_{t} - \mathrm{H}_{1/\sigma}\left(\mathbf{y}_{t}/\sigma +  \hat{\mathbf{m}}_{t}\right)$}

\State {$\mathbf{m}_{t+1} \longleftarrow \text{prox}_{\tau f}(\mathbf{m}_{t} - \tau\mathbf{y}_{t+1})$}

\State{$\hat{\mathbf{m}}_{t+1} \longleftarrow \mathbf{m}_{t+1} + \theta (\mathbf{m}_{t+1} - \mathbf{m}_{t})$}

\State{where $\tau$ is the step size, $1/\sigma$ is the noise level for the CNN-based denoiser, and $\theta$ is the acceleration parameter.}

\EndFor
\end{algorithmic}
\end{algorithm}
%
\subsection{Variational inference and its regularization formulation}\label{sec:theory-vi}
In this section, we introduce the variational inference framework for probabilistic seismic inversion. Compared to the deterministic inversion introduced in the previous section, the variational inference approach approximates the posterior distribution through a surrogate distribution $q(\mathbf{m})$ that is easy to sample (e.g., Gaussian distribution). This approach frames the computationally costly sampling procedure, such as the one required by MCMC algorithms, into an optimization problem in the space of distributions, i.e., a functional minimization problem of the KL divergence between the surrogate $q(\mathbf{m})$ and a target posterior distributions $\pi(\mathbf{m})$~\cite{blei2017vi, zhang2018vi}. The KL divergence is given by
\begin{equation} \label{eq:kl-div}
    \mathrm{KL}(q(\mathbf{m})||\pi(\mathbf{m})) = \mathrm{E}_{\mathbf{m} \sim q}[-\log\pi(\mathbf{m}) + \log q(\mathbf{m})],
\end{equation}
where $\mathbf{m}$ in the right hand side of ~\eqref{eq:kl-div} above is 
obtained by sampling the surrogate distribution $q(\mathbf{m})$, over which we evaluate the expectation. The variational inference approximates a target posterior distribution $\pi(\mathbf{m})$ using $q(\mathbf{m})$ by minimizing the equation \eqref{eq:kl-div}, that is,
\begin{equation} \label{eq:min-kl-vi}
\begin{split}
    q^{\ast} &= \argmin_{q} \mathrm{E}_{\mathbf{m} \sim q}[-\log\pi(\mathbf{m}) + \log q(\mathbf{m})]\\
    &= \argmin_{q} \mathrm{KL}(q(\mathbf{m})||\pi(\mathbf{m}))
\end{split}
\end{equation}
where the choice of $q(\mathbf{m})$ is crucial and gives rise to different variational inference methods. According to Liu and Wang~\cite{liu2016svgd}, the best choice of $q(\mathbf{m})$ should strike a balance between three criteria: i) \emph{accuracy:} $q(\mathbf{m})$ should be broad enough to accurately approximate a large class of target distributions, ii) \emph{tractability:} $q(\mathbf{m})$ should be easy to sample (e.g., Gaussian distribution), and iii) \emph{solvability:} The choice of $q(\mathbf{m})$ should be simple so that the minimization problem in equation \eqref{eq:min-kl-vi} can be solved efficiently. This \emph{unconstrained variational inference} framework considers the prior information about the solutions $\mathbf{m}$ only through the prior distribution in Bayes' theorem, as presented in equation~\eqref{eq:unnormalized-bayes}.

Prior knowledge about the solution $\mathbf{m}$ can be further imposed on the posterior distribution by introducing a regularization term into equation~\eqref{eq:min-kl-vi}. By doing so, we restrict our posterior distribution further by only providing samples that belong in the regularized posterior distribution (i.e., constrained space of the posterior distribution). We call such an approach the \emph{regularized variational inference} framework~\cite{altun2006reg, zhu2014reg}. The \emph{regularized variational inference} framework can be naturally formulated as:
\begin{equation} \label{eq:reg-bayes}
\begin{split}
    q^{\ast} &= \argmin_{q} \mathrm{E}_{\mathbf{m} \sim q}[-\log\pi(\mathbf{m}) + \log q(\mathbf{m})] + \mathcal{R}_{q(\mathbf{m})}(\mathbf{m})\\
    &= \argmin_{q} \mathrm{KL}(q(\mathbf{m})||\pi(\mathbf{m})) + \mathcal{R}_{q(\mathbf{m})}(\mathbf{m})
\end{split}
\end{equation}
where $\mathcal{R}_{q(\cdot)}(\cdot)$ is a user-defined regularization function acting on the samples $\{\mathbf{m}_{i}\}_{i=1}^{N}$ drawn from $q(\mathbf{m})$. In the machine learning community, this formulation has been utilized by Sinha and Dieng~\cite{sinha2021ae} to regularize the training process of variational autoencoders (VAEs) using the so-called consistency regularization to enforce consistency in the VAE latent representation as well as by the authors of~\cite{liu2021svgd, zhang2022svgd} to impose trustworthy-related constraints onto machine learning systems in order to ensure safety. In geophysical inversion applications, such regularization is needed not only for uncertainty quantification purposes but also a way to encourage a better representation of posterior samples, such as high-resolution samples representative of the subsurface structures, which can be used for post-inference tasks, e.g., history matching~\cite{oliver2008histmatch}.

In this section, we have introduced the unconstrained and constrained variational inference frameworks, whose numerical optimization process can be accomplished using several approaches, e.g., variational Bayes~\cite{urozayev2022viseismic}, deep neural network representation~\cite[e.g.,][]{baptista2020vi, kruse2021vi, siahkoohi2021vi, siahkoohi2022vi} or particle evolution update such as SVGD~\cite[e.g.,][]{liu2016svgd, wang2019msvgd, liu2021svgd, shi2021svgd, zhang2022svgd}. In this work, we propose to solve the variational inference through SVGD, which will be discussed in detail next.
%
\subsubsection{Stein Variational Gradient Descent (SVGD)} \label{sec:theory-svgd}
We provide an overview of the SVGD algorithm and propose our PnP-SVGD algorithm as a novel algorithm for sampling from a regularized target posterior distribution. The SVGD algorithm was proposed by \cite{liu2016svgd} in 2016 as a general-purpose variational inference algorithm based on particle evolution. It starts from an arbitrary initial density $q_{0}$ and the associated set of particles $\mathbf{m}_{i, 0} \sim q_{0}$, where $\mathbf{m}_{i, 0}$ is a $d$-dimensional vector containing the parameters of interest. These particles are updated through a smooth transform $\mathbf{T}(\mathbf{m})=\mathbf{m} + \eta\phi(\mathbf{m})$, where $\phi(\mathbf{m})$ is a smooth function that characterizes the perturbation direction and the scalar $\eta$ acts as the perturbation magnitude. We assume the transformation $\mathbf{T}$ is invertible and define $q_{[\mathbf{T}]}(\mathbf{m})$ as the transformed density by the transformation $\mathbf{T}$. By using operators based on Stein's method, the gradient of the KL divergence between $q_{[\mathbf{T}]}(\mathbf{m})$ and a target posterior $\pi(\mathbf{m})$ with respect to the perturbation magnitude $\eta$ can be described through the \emph{Stein identity} as:
\begin{equation} \label{eq:grad-kl}
    \nabla_{\varepsilon}\mathrm{KL}(q_{[\mathbf{T}]}(\mathbf{m})||\pi(\mathbf{m}))|_{\varepsilon=0} = -\mathrm{E}_{\mathbf{m} \sim q}[\text{trace}(\mathcal{A}_{\pi}\phi(\mathbf{m}))],
\end{equation}
where $\mathcal{A}_{\pi}\phi(\mathbf{m}) = \nabla_{\mathbf{m}}\log\pi(\mathbf{m})\phi(\mathbf{m}) + \nabla_{\mathbf{m}}\phi(\mathbf{m})$ is the \emph{Stein operator} \cite{stein1972stein, gorham2015stein, gorham2017stein, gorham2019stein, izzatullah2020stein}. Equation \eqref{eq:grad-kl} provides a form of functional gradient descent direction of the KL divergence. Therefore, the KL divergence can be minimized in an iterative manner by a small magnitude of $\eta$. To minimize the KL divergence as fast as possible, the function $\phi(\mathbf{m})$ should be chosen from a set of functions such that the decreasing rate of $-\frac{d}{dt}(q_{[\mathbf{T}]}(\mathbf{m})||\pi(\mathbf{m}))$ is maximized. Commonly, this set of functions is chosen to be positive definite with continuously differentiable kernel functions from the \emph{reproducing kernel Hilbert space (RKHS)}~\cite{berlinet2011rkhs, paulsen2016rkhs}. By the reproducing property of RKHS and Stein identity in equation \eqref{eq:grad-kl}, we can show that the optimal $\phi^{\ast}$ is
\begin{equation}\label{eq:optimal-phi}
    \phi^{\ast}(\cdot) = \mathrm{E}_{\mathbf{m} \sim q}[\nabla_{\mathbf{m}}\log\pi(\mathbf{m})k(\mathbf{m}, \cdot) + \nabla_{\mathbf{m}}k(\mathbf{m}, \cdot)],
\end{equation}
where $k(\mathbf{m}, \cdot)$ is a RKHS function. Through equation \ref{eq:optimal-phi}, the SVGD of \cite{liu2016svgd} can be derived as
\begin{equation}\label{eq:svgd}
\begin{split}
    \mathbf{m}_{i, t+1} &= \mathbf{m}_{i, t} - \eta_{t}\mathrm{E}_{\mathbf{m} \sim q_{t}}[\nabla_{\mathbf{m}_{i, t}}\log\pi(\mathbf{m}_{i, t})k(\mathbf{m}_{i, t}, \mathbf{m}) + \nabla_{\mathbf{m}_{i, t}}k(\mathbf{m}_{i, t}, \mathbf{m})]\\
    &= \mathbf{m}_{i, t} - \frac{\eta_{t}}{N}\sum^{N}_{i=1}[\nabla_{\mathbf{m}_{i, t}}\log\pi(\mathbf{m}_{i, t})k(\mathbf{m}_{i, t}, \mathbf{m}) + \nabla_{\mathbf{m}_{i, t}}k(\mathbf{m}_{i, t}, \mathbf{m})]\\
    &= \mathbf{m}_{i, t} - \eta_{t}\phi^{\ast}_{t}(\mathbf{m}_{i, t}),
\end{split}
\end{equation}
for $N$ particles $\{\mathbf{m}_{i,t}\}_{i=1}^{N} \in \mathrm{R}^{d}$. We summarize the SVGD procedure in Algorithm~\ref{alg:svgd}. SVGD forms a natural counterpart of gradient descent for optimization by iteratively updating those particles from their initial density $q_{0}(\mathbf{m})$ to match the target posterior $\pi(\mathbf{m})$ by minimizing the KL divergence. Also, SVGD has found great success in approximating unconstrained distributions for probabilistic deep learning \cite{feng2017svgd, haarnoja2017svgd, yoon2018svgd} as well as for Bayesian geophysical inversion \cite{zhang2020svgdseismic, zhang2021svgdseismic, ramgraber2021svgd, smith2022svgd}.
%
\begin{algorithm}
\caption{Stein Variational Gradient Descent (SVGD)}\label{alg:svgd}
\textbf{Input:} A target posterior distribution $\pi(\mathbf{m})$ and a set of initial particles $\{\mathbf{m}_{i,t}\}_{i=1}^{N}$.
\textbf{Output:} A set of particles $\{\mathbf{m}_{i,t}\}_{i=1}^{N}$ that approximates the target posterior distribution.
\begin{algorithmic}
\For{iteration $t$} 

\State {$\mathbf{m}_{i,t+1} \longleftarrow \mathbf{m}_{i,t} - \eta_{t}\phi^{\ast}_{t}(\mathbf{m}_{i, t})$ where $\phi^{\ast}(\mathbf{m}) = \frac{1}{N}\sum^{N}_{i=1}[\nabla_{\mathbf{m}_{i, t}}\log\pi(\mathbf{m}_{i, t})k(\mathbf{m}_{i, t}, \mathbf{m}) + \nabla_{\mathbf{m}_{i, t}}k(\mathbf{m}_{i, t}, \mathbf{m})]$}

\State{where $\eta_{t}$ is the step size at the $t$-th iteration.}

\EndFor
\end{algorithmic}
\end{algorithm}

\subsubsection{Plug-and-Play Stein Variational Gradient Descent (PnP-SVGD)} \label{sec:theory-pnp-svgd}
In many cases, we would like to impose user-defined constraints or regularizations in addition to approximating $\pi(\mathbf{m})$ to better represent the posterior samples. In geophysical applications, for example, high-resolution samples representing the subsurface structures are required in performing a reliable reservoir history matching procedure. This objective can be achieved by imposing constraints or regularizations directly onto the posterior distribution. To sample from the regularized posterior distribution, we consider solving the following minimization problem 
\begin{equation} \label{eq:reg-bayes-pnp}
\begin{split}
    q^{\ast} &= \argmin_{q} \mathrm{KL}(q(\mathbf{m})||\pi(\mathbf{m})) + \mathcal{R}_{q(\mathbf{m})}(\mathbf{m})\\
    &= \argmin_{q} \mathcal{L}(q(\mathbf{m})) + \mathcal{R}_{q(\mathbf{m})}(\mathbf{m}).
\end{split}
\end{equation}
An interesting observation is that the above problem resembles the convex optimization problem as formulated in equation~\eqref{eq:pnp-fg}. Here, we can leverage a typical convex optimization method in solving such a problem by evaluating the proximal operator in equation~\eqref{eq:pnp-prox}. This can be performed by taking the regularization functional $\mathcal{R}_{q(\mathbf{m})}(\mathbf{m})$ as the function that needs to be evaluated in the proximal operator. By leveraging this formulation, we proposed an algorithm based on the forward-backward splitting (FBS), which can be described as
\begin{equation}\label{eq:fbs-svgd}
\begin{split}
    \mathbf{m}_{i, t+1} &= \mathrm{H}_{\sigma}\big[\mathbf{m}_{i, t} - \eta_{t}\mathrm{E}_{\mathbf{m} \sim q_{t}}[\nabla_{\mathbf{m}_{i, t}}\log\pi(\mathbf{m}_{i, t})k(\mathbf{m}_{i, t}, \mathbf{m}) + \nabla_{\mathbf{m}_{i, t}}k(\mathbf{m}_{i, t}, \mathbf{m})]\big]\\
    &= \mathrm{H}_{\sigma}\big[\mathbf{m}_{i, t} - \frac{\eta_{t}}{N}\sum^{N}_{i=1}[\nabla_{\mathbf{m}_{i, t}}\log\pi(\mathbf{m}_{i, t})k(\mathbf{m}_{i, t}, \mathbf{m}) + \nabla_{\mathbf{m}_{i, t}}k(\mathbf{m}_{i, t}, \mathbf{m})]\big]\\
    &= \mathrm{H}_{\sigma}\big[\mathbf{m}_{i, t} - \eta_{t}\phi^{\ast}_{t}(\mathbf{m}_{i, t})\big],
\end{split}
\end{equation}
where $\mathrm{H}_{\sigma}$ is a denoiser with $\sigma \geq 0$ as a noise parameter acting on the samples $\{\mathbf{m}_{i, t}\}_{i=1}^{N}$ drawn from $q_{t}(\mathbf{m})$. We call the algorithm in equation \eqref{eq:fbs-svgd} \emph{Plug-and-Play Stein Variational Gradient Descent (PnP-SVGD)}–a novel sampling algorithm for sampling from a regularized target posterior distribution. Generally, the PnP framework can be formulated for many proximal algorithms in a deterministic setting~\cite{meinhardt2017pnp}, such as FBS, ADMM, and PD. However, we favor the FBS formulation for PnP-SVGD because of its close connection with the proximal LMC algorithm, which was developed based on the FBS formulation~\cite{pereyra2016lmc, salim2019lmc}. Also, SVGD and LMC are derived from the same Langevin diffusion, which shares a common theoretical background~\cite[sec.~2.1]{liu2021svgd}. Furthermore, the FBS formulation simplifies the SVGD formulation for probabilistic inference compared to the others and provides a natural connection to its deterministic counterpart. Through PnP-SVGD, we pursue the high-resolution samples representing the subsurface structures by regularizing the approximation of the target posterior distributions and quantifying the model uncertainty from the obtained samples. The PnP-SVGD procedure is summarized in Algorithm~\ref{alg:pnp-svgd}.
%
\begin{algorithm}
\caption{PnP-Stein Variational Gradient Descent (PnP-SVGD)}\label{alg:pnp-svgd}
\textbf{Input:} A target posterior distribution $\pi(\mathbf{m})$, a CNN-Based denoiser $\mathrm{H}_{\sigma}$, and a set of initial particles $\{\mathbf{m}_{i,t}\}_{i=1}^{N}$.
\textbf{Output:} A set of particles $\{\mathbf{m}_{i,t}\}_{i=1}^{N}$ that approximates the target posterior distribution.

\begin{algorithmic}
\For{iteration $t$} 

\State {$\mathbf{m}_{i,t+1} \longleftarrow \mathbf{m}_{i,t} - \eta_{t}\phi^{\ast}_{t}(\mathbf{m}_{i, t})$ where $\phi^{\ast}(\mathbf{m}) = \frac{1}{N}\sum^{N}_{i=1}[\nabla_{\mathbf{m}_{i, t}}\log\pi(\mathbf{m}_{i, t})k(\mathbf{m}_{i, t}, \mathbf{m}) + \nabla_{\mathbf{m}_{i, t}}k(\mathbf{m}_{i, t}, \mathbf{m})]$}

\State{$\mathbf{m}_{i,t+1} \longleftarrow \mathrm{H}_{\sigma}(\mathbf{m}_{i,t+1})$}

\State{where $\eta_{t}$ is the step size at the $t$-th iteration and $\sigma$ is the noise level for the denoiser.}

\EndFor
\end{algorithmic}
\end{algorithm}

where we have decided $\mathrm{H}_{\sigma}$ to be a deep learning-based denoiser such as DnCNN~\cite{zhang2017pnp} or DRUNet~\cite{zhang2021pnp}; however, we note that the very same theory applies for any other choice of denoiser. 
\section{Numerical Examples}
This section demonstrates the proposed methodology through a number of numerical examples of post-stack seismic inversion applied to both a realistic synthetic dataset and a field dataset from the Volve oilfield. We highlight the potentials of our PnP-SVGD method over vanilla SVGD on the synthetic dataset, then focus only on the PnP-SVGD for the Volve dataset. For both algorithms, we use the RBF kernel with bandwidth chosen by the standard median trick, that is, we use $k_{t}(\mathbf{m}, \mathbf{m}^{'}) = \exp({-\|\mathbf{m} - \mathbf{m}^{'}\|^{2} / \sigma_{t}^{2}})$, where the bandwidth $\sigma_{t}$ is set by $\sigma_{t} = \textbf{Median}\{\|\mathbf{m}_{i, t} - \mathbf{m}_{j,t}\|:i \neq j\}$ based on the particles at the $t$-th iteration. Furthermore, we also compare our posterior samples' mean with the inversion result from the PnP-PD of~\cite{romero2022pnp} both in terms of quality and resolution by means of the signal-to-noise ratio (SNR).

We consider sampling from the following log-posterior model for the model uncertainty quantification task in all numerical examples:
\begin{equation}\label{eq:log-posterior}
    \log \pi(\mathbf{m}) = \underbrace{-\frac{1}{2}\|\mathbf{G}\mathbf{m} - \mathbf{d}\|^{2}_{\mathbf{C}_{\mathbf{d}}}}_{\log p(\mathbf{d}|\mathbf{m})}  \underbrace{-\frac{1}{2}\|\mathbf{m}\|^{2}_{\mathbf{C}_{\mathbf{m}}} - \|\mathbf{D}\mathbf{m}\|_{1}}_{\sum^{2}_{i=1} \log p_{i}(\mathbf{m})} - \underbrace{\log Z}_{\text{Const.}},
\end{equation}
where $\|\mathbf{x}\|^{2}_{\mathbf{A}} = \mathbf{x}^{T}\mathbf{A}\mathbf{x}$. The first term on the right hand side of equation\eqref{eq:log-posterior} represents the log-likelihood term which describes the conditional probability density for the unknown AI model $\mathbf{m}$ given the observed seismic data $\mathbf{d}$ with $\mathbf{C}_{\mathbf{d}} = \sigma_{\mathbf{d}} \mathbf{I}$ as the noise covariance matrix. For both numerical examples we fixed $\sigma_{\mathbf{d}} = 10^{-2}$. The second term represents the composition of two model log-priors, which are equivalent to the Tikhonov and anisotropic total-variation (TV) regularization terms. For both $\mathbf{C}_{\mathbf{m}}$ and $\mathbf{D}$ in the log-priors terms are taken to be the directional derivatives matrices. The constant term $\log Z$ in the above equation can be ignored without affecting the inference procedure. For the PnP-PD and PnP-SVGD, we use the DRUNet from~\cite{zhang2021pnp} as the deep denoiser in both the optimization and sampling procedures because of its superior performance over other CNN-based denoisers. As argued in~\cite{romero2022pnp}, this could be attributed to: i) its bias-free formulation (locally homogeneous denoising map, which is invariant to scaling), ii) the extra input channel, which represents the noise level map chosen to match the value defined by the PD algorithm for the denoising step, and iii) the effectiveness of UNet for image-to-image translation, the bigger modeling capacity arising from the ResNet blocks, and the versatility of FFDNet to handle various noise ranges. All the numerical examples have been implemented using PyTorch~\cite{paszke2017} and PyLops~\cite{ravasi2020}.

\subsection{Synthetic data example: Hess model}
\begin{figure*}[!htb]
  \centering
  \includegraphics[width=\textwidth]{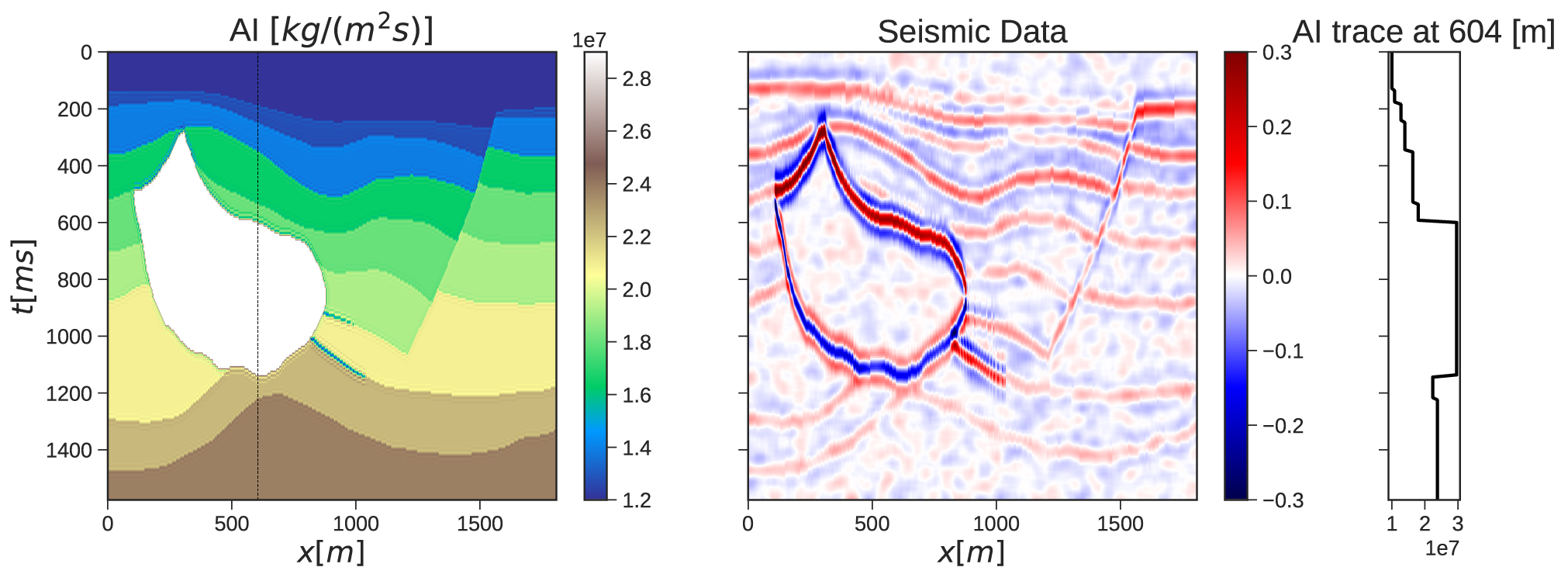}
  \caption{Hess model. \textbf{Left:} Acoustic impedance model. \textbf{Center:} seismic data. \textbf{Right:} acoustic impedance trace at $x=604$m.}
  \label{fig:3-1-Hess-data}
\end{figure*}

\begin{figure*}[!htb]
  \centering
  \includegraphics[width=\textwidth]{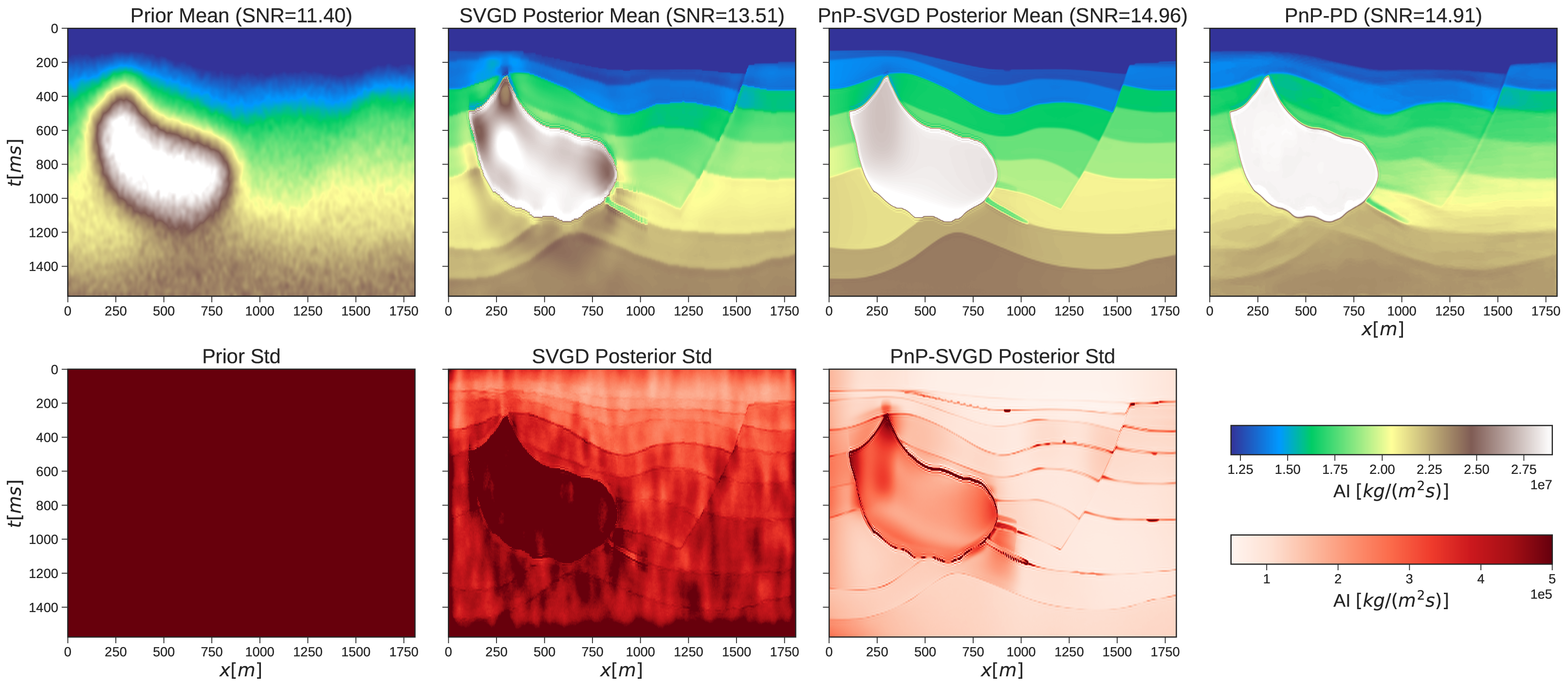}
  \caption{Hess model statistics. \textbf{Top row:} Mean models of the samples drawn from prior and posterior distributions, along with inverted model from the deterministic PnP-PD optimization procedure. \textbf{Bottom row:} The samples standard deviation of the samples drawn from prior and posterior distributions.}
  \label{fig:3-1-Hess-models}
\end{figure*}

The proposed PnP-SVGD algorithm is here compared to the vanilla SVGD for post-stack seismic inversion applied on synthetic seismic data (Figure~\ref{fig:3-1-Hess-data}: Center) computed using the post-stack modeling operator $\mathbf{G}$ as described in Section~\ref{sec:theory-bayes} on the Hess model illustrated in the left of Figure~\ref{fig:3-1-Hess-data}. Following~\cite{ravasi2022}, the observed seismic data is generated using a Ricker wavelet with a peak frequency of $8$ Hz. Furthermore, to mimic the nature of field seismic data, we add band-passed noise created by filtering Gaussian noise along both the vertical and horizontal axes to the synthetic observed data. We sample $100$ particles (samples) from an initial distribution, which here we consider to be Gaussian distribution with a smooth AI background model as the mean and a diagonal covariance matrix $\mathbf{C}_{q_{0}} = 0.5 \mathbf{I}$. For both algorithms, we perform $50$ iterations and we collect all the $100$ updated particles as the posterior samples for statistical analysis at the end of simulations. We also perform $100$ optimization iterations using the PnP-PD algorithm for posterior samples mean model comparison. We illustrate the statistics for this Hess model simulations in Figure~\ref{fig:3-1-Hess-models}.

\begin{figure*}[!htb]
  \centering
  \includegraphics[width=\textwidth]{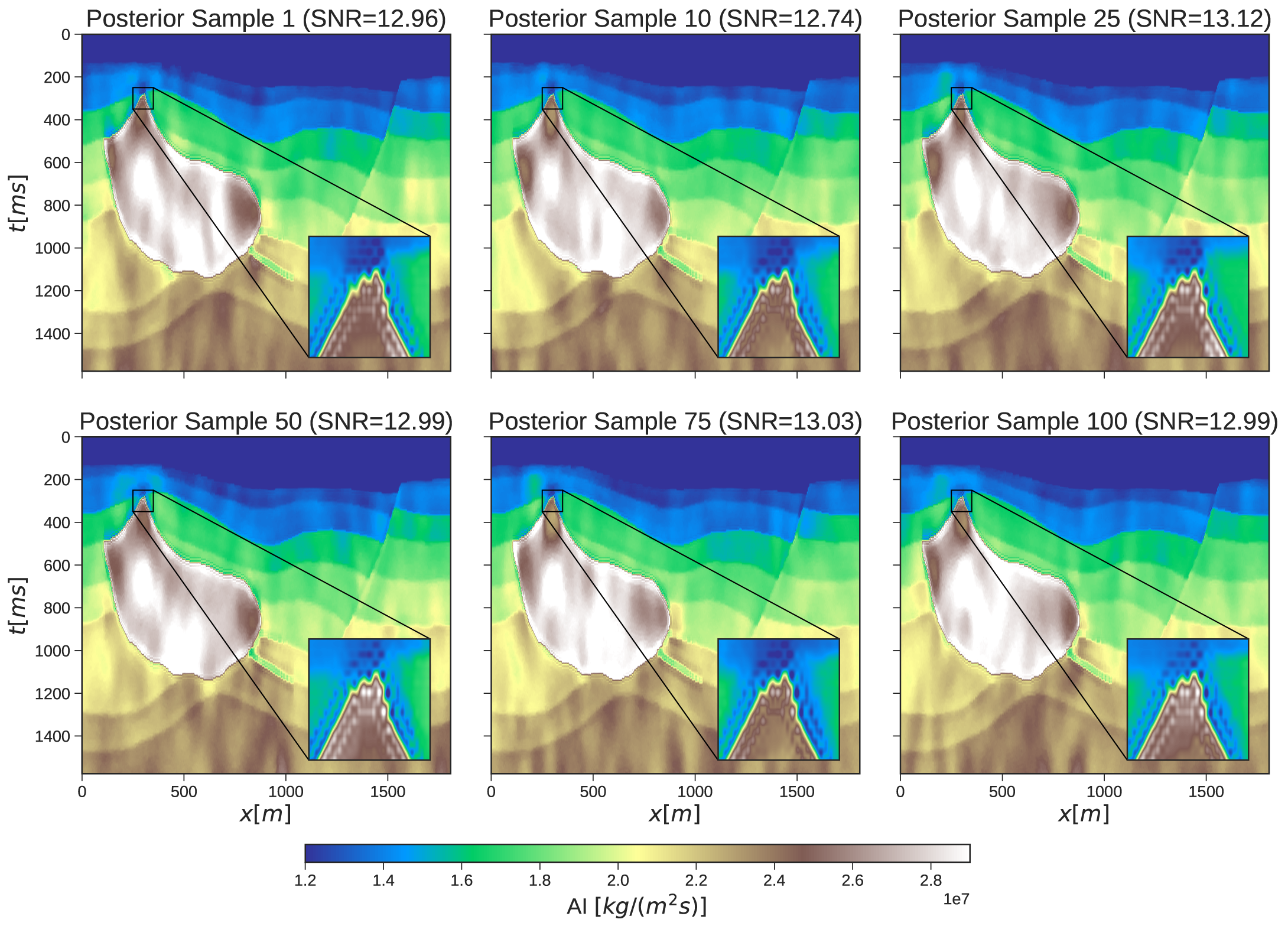}
  \caption{Hess model posterior samples from the SVGD procedure.}
  \label{fig:3-1-Hess-posterior-samples-svgd}
\end{figure*}

\begin{landscape}
\begin{figure*}[!htb]
  \vspace{11\baselineskip}
  \includegraphics[width=1.3\textwidth, keepaspectratio]{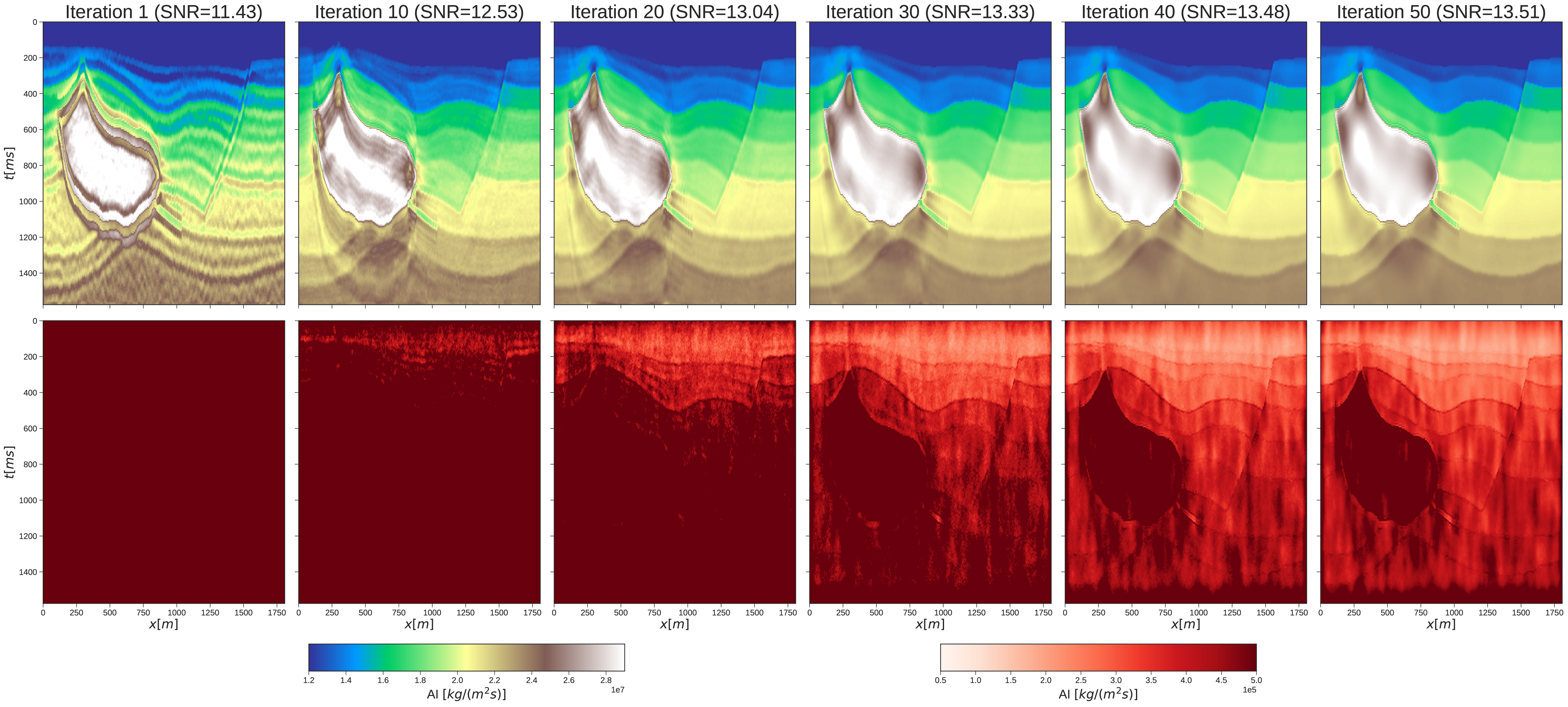}
  \caption{Hess model samples mean and standard deviation evolution per iteration for the SVGD procedure.}
  \label{fig:3-1-Hess-evo-svgd}
\end{figure*}
\end{landscape}

\begin{figure*}[!htb]
  \centering
  \includegraphics[width=\textwidth]{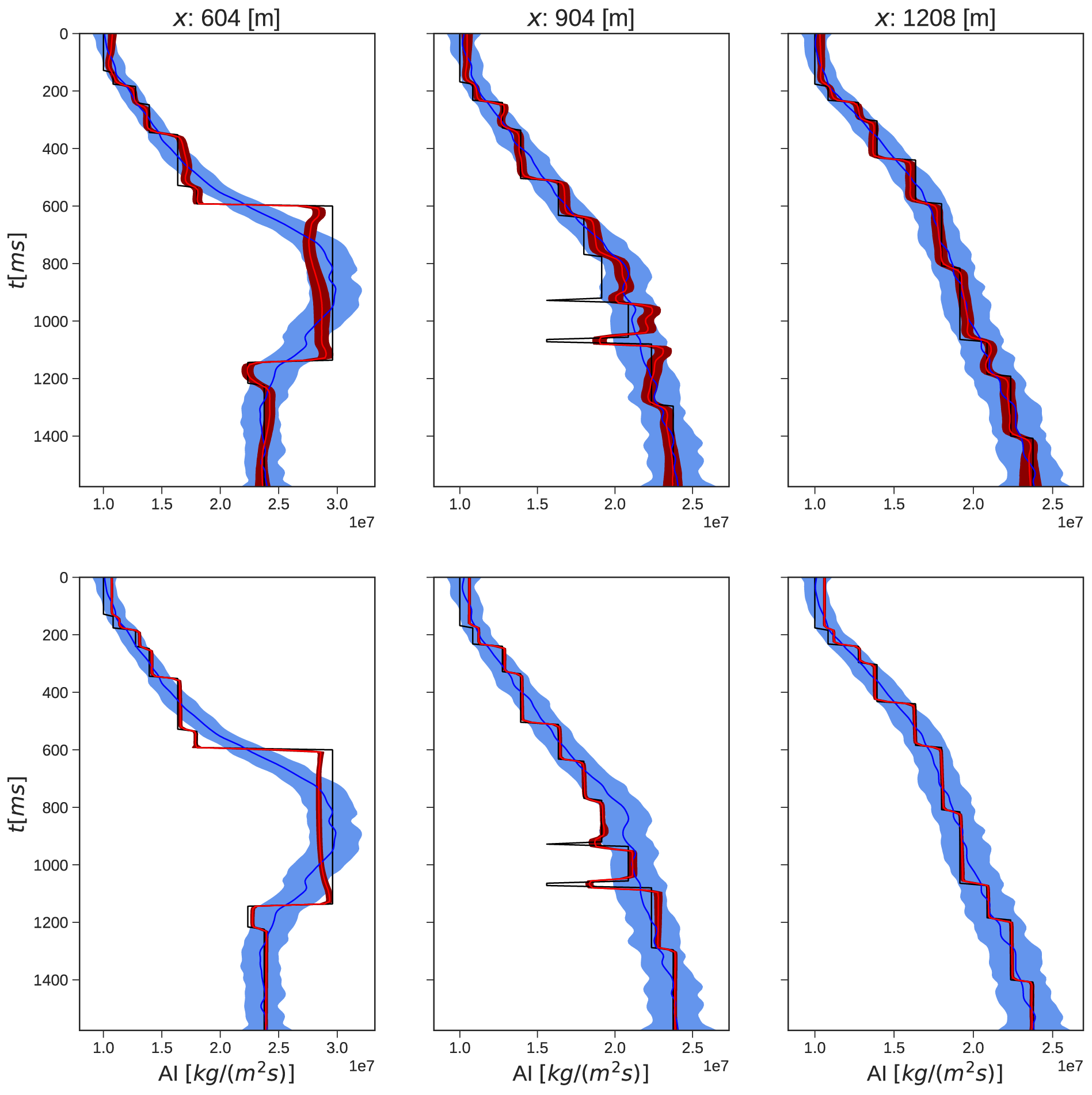}
  \caption{Hess model statistics per trace. \textbf{Top row:} SVGD. \textbf{Bottom row:} PnP-SVGD. The blue and red curves and regions represent the samples mean and 1-$\sigma$ credible-interval for the prior and posterior distributions, respectively.}
  \label{fig:3-1-Hess-traces}
\end{figure*}

As expected, SVGD is unable to retrieve the high-frequency components of the subsurface, yielding a low-resolution posterior mean model with SNR of $13.51$ dB. Although a relatively good posterior mean model (i.e., featuring the subsurface structures relatively well), the posterior samples from SVGD suffer from noise imprints, as shown in Figure~\ref{fig:3-1-Hess-posterior-samples-svgd}, due to the ill-posedness of seismic inversion where posterior samples are affected by the noise present in the observed data, leading to an unstable inference procedure in the context of SVGD. However, such posterior samples are expected from SVGD because it explores the posterior distribution and provides us only with posterior samples that respect the likelihood without constraining the resolution of the samples. Thus, it leads to high uncertainty in the inversion results, as illustrated in the bottom row of Figure~\ref{fig:3-1-Hess-models} and in the top row of Figure~\ref{fig:3-1-Hess-traces} for per trace statistics. We also present the evolution of the SVGD posterior samples mean and standard deviation for selected iterations in Figure~\ref{fig:3-1-Hess-evo-svgd} where we observed that SVGD is unable to further reduce the model uncertainty and the noise imprints as iteration increases. 

\begin{figure*}[!htb]
  \centering
  \includegraphics[width=\textwidth]{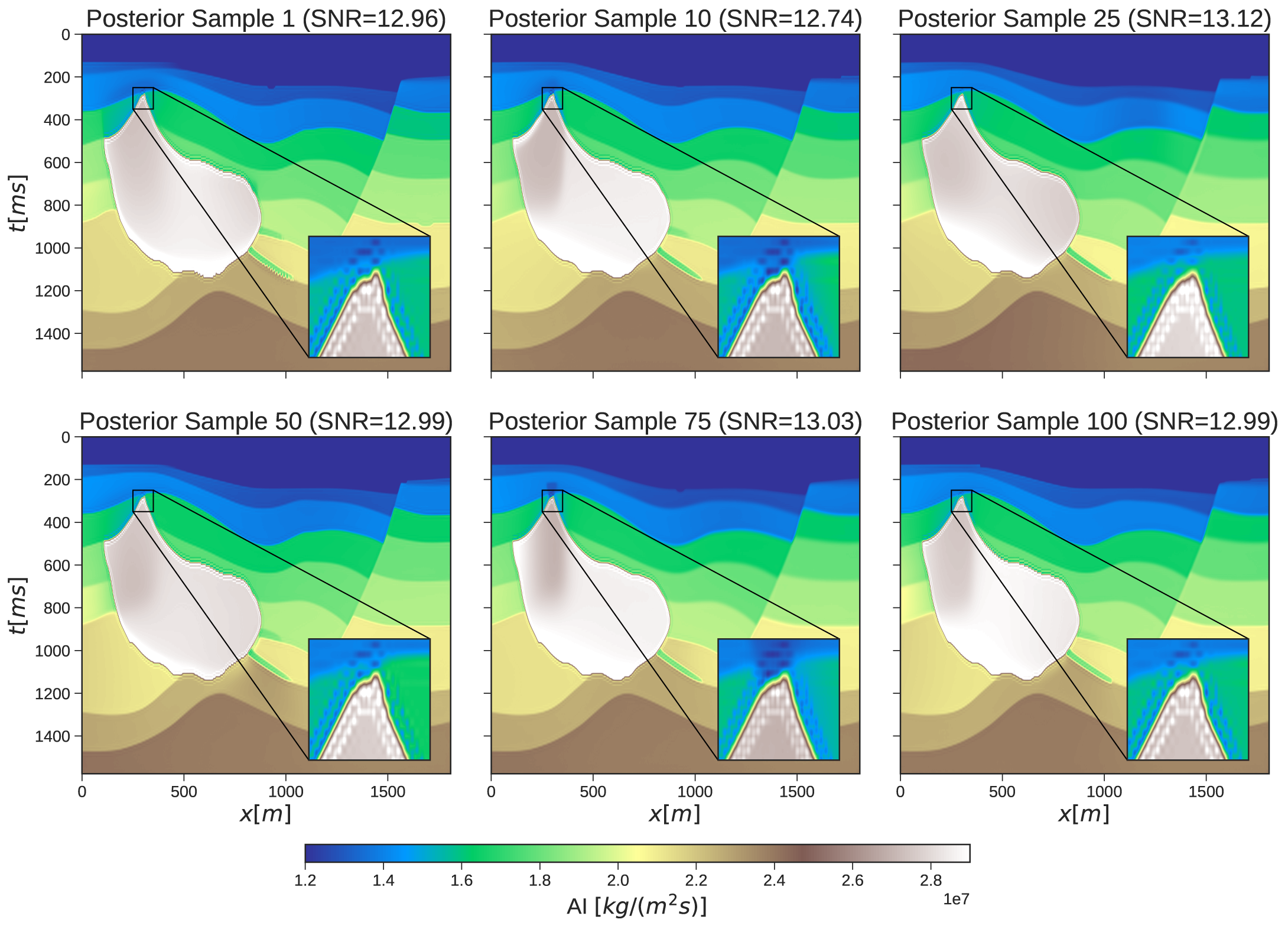}
  \caption{Hess model posterior samples from the PnP-SVGD procedure.}
  \label{fig:3-1-Hess-posterior-samples}
\end{figure*}

\begin{landscape}
\begin{figure*}[!htb]
  \vspace{11\baselineskip}
  \includegraphics[width=1.3\textwidth, keepaspectratio]{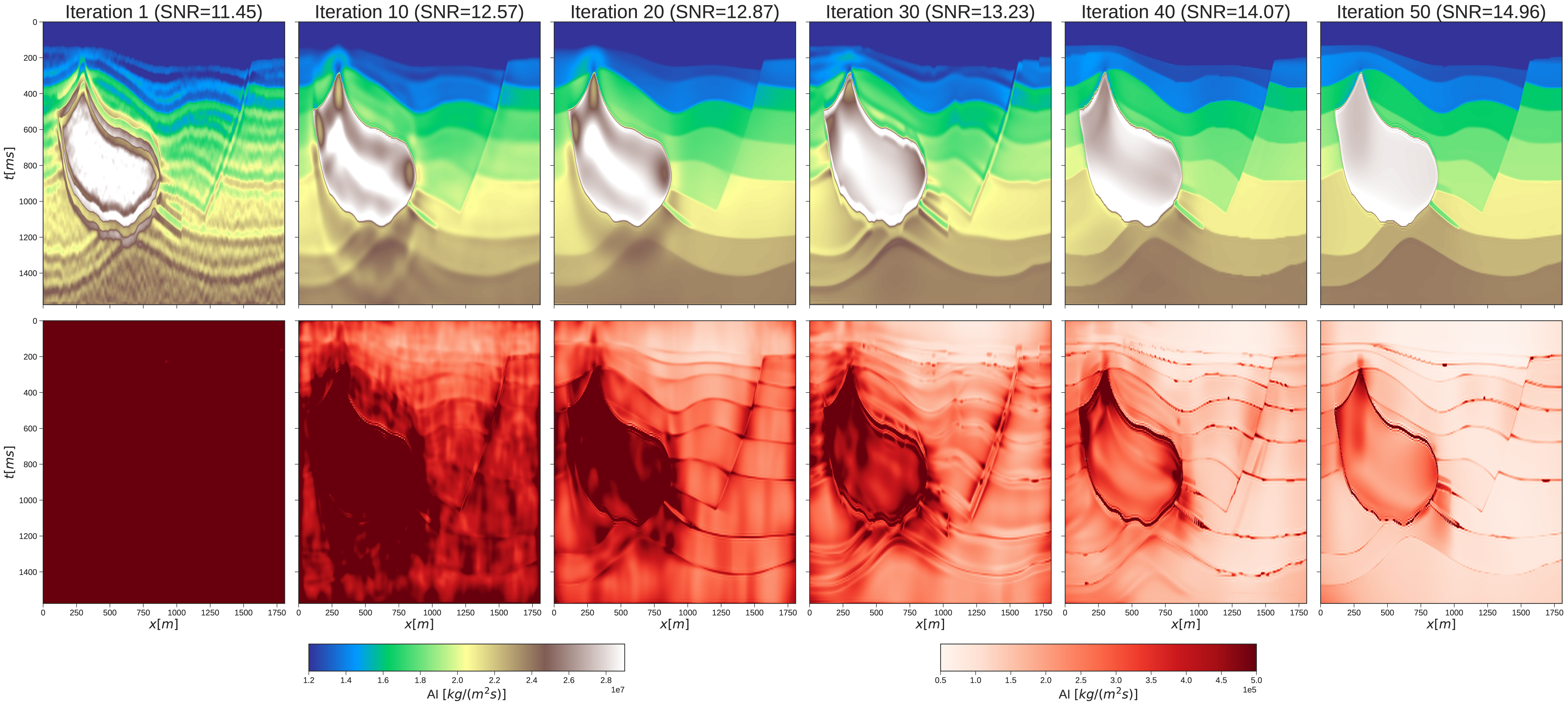}
  \caption{Hess model samples mean and standard deviation evolution per iteration for the PnP-SVGD procedure.}
  \label{fig:3-1-Hess-evo}
\end{figure*}
\end{landscape}

\begin{figure*}[!htb]
  \centering
  \includegraphics[width=0.85\textwidth]{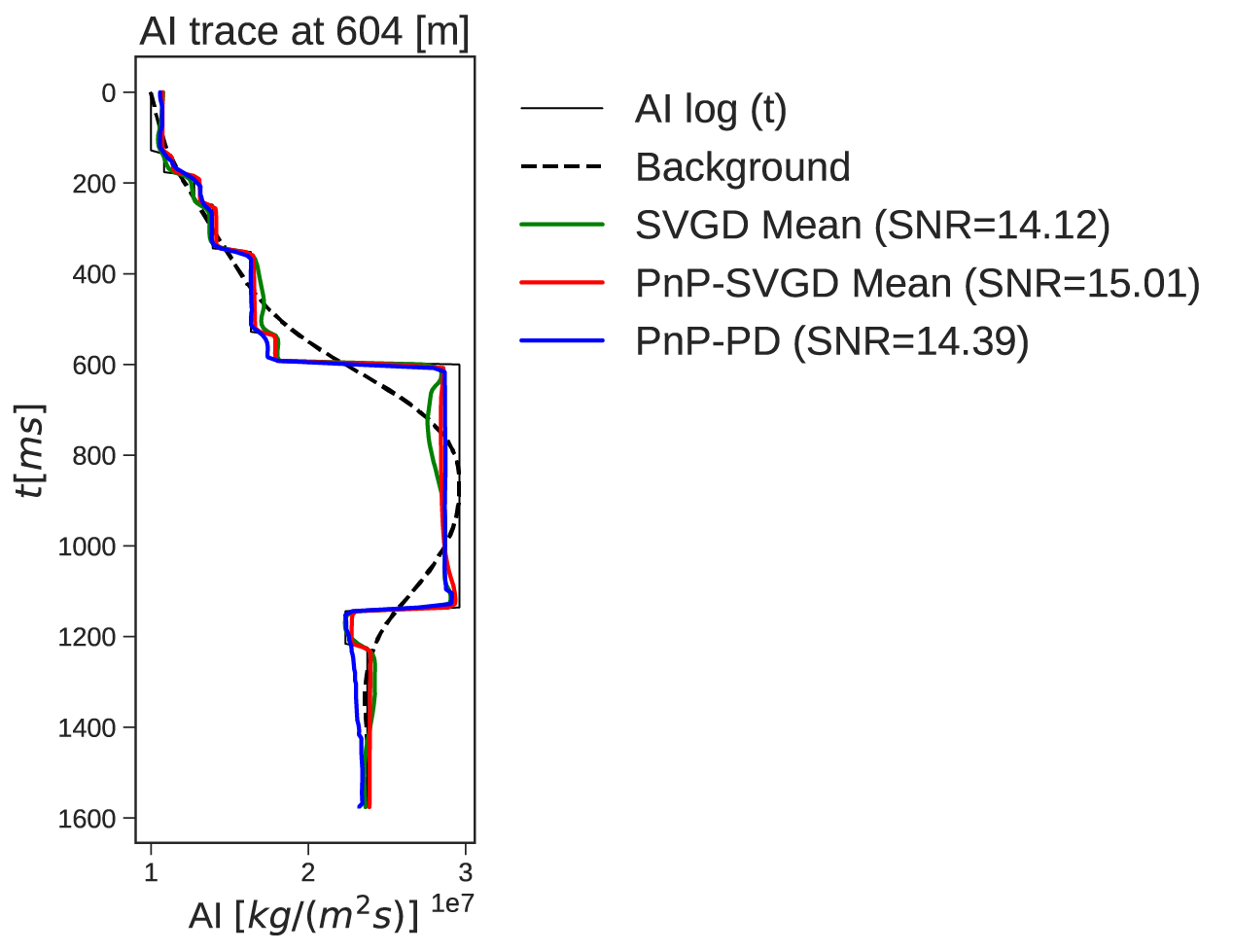}
  \caption{Hess model acoustic impedance trace comparison at $604$m.}
  \label{fig:3-1-Hess-log}
\end{figure*}

PnP-SVGD, on the other hand, can retrieve high-resolution posterior samples as confirmed by an overall SNR of $14.96$ dB for the posterior samples mean model in Figure~\ref{fig:3-1-Hess-models} and our results in Figure~\ref{fig:3-1-Hess-posterior-samples}. We attribute such an improvement to the additional regularization term implemented via a DRUNet denoiser, which penalizes noisy features in the samples that do not respect our expectation of the subsurface. In other words, by reformulating SVGD in the PnP framework, we leverage the powerful regularization effect from DRUNet in driving the posterior sampling procedure and constraining our posterior towards the region of the high-resolution samples, yielding low model uncertainty as observed in the bottom row of Figure~\ref{fig:3-1-Hess-models} and the top row of Figure~\ref{fig:3-1-Hess-traces} for per trace statistics. Furthermore, PnP-SVGD is robust with respect to noise present in the observed data and provides us with a stable and fast inference procedure compared to vanilla SVGD, as confirmed by Figure~\ref{fig:3-1-Hess-evo}. Finally, in comparison to the PnP-PD, PnP-SVGD records a higher SNR than PnP-PD and visually outperforms the PnP-PD solution in terms of image resolution, especially in the deeper region of the AI model, as confirmed by the AI trace at $604$ m shown in Figure~\ref{fig:3-1-Hess-log}.

\begin{figure*}[!htb]
  \centering
  \includegraphics[width=\textwidth]{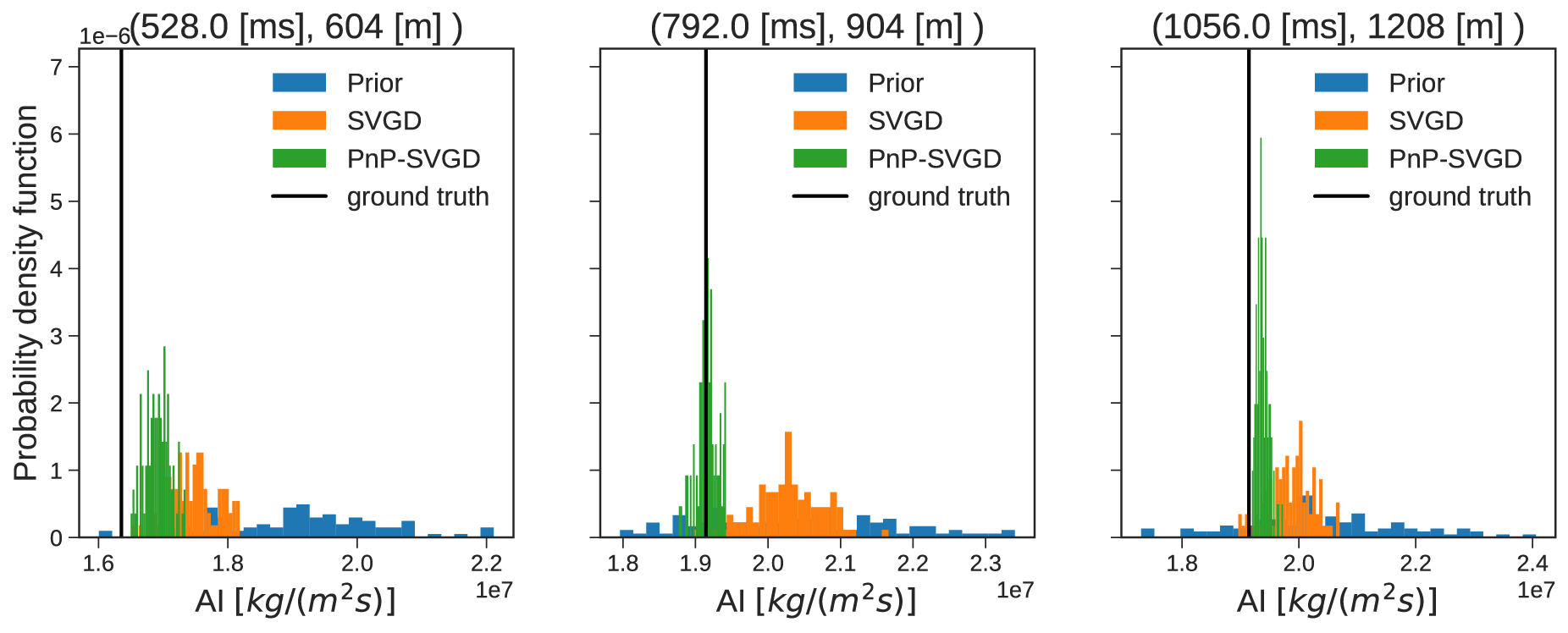}
  \caption{Hess model pointwise histogram for points located at \textbf{left:} ($528$ [ms], $604$ [m]), \textbf{center:} ($792$ [ms], $904$ [m]), and \textbf{right:} ($1056$ [ms], $1208$ [m]).}
  \label{fig:3-1-Hess-hist}
\end{figure*}

To illustrate how the posterior for both algorithms are distributed, we also calculated pointwise histograms at three locations, as illustrated in Figure~\ref{fig:3-1-Hess-hist}. Histograms from the prior are calculated by randomly sampling from the initial distribution of $q_{0}$. As expected, the histograms for the posterior are considerably narrower than prior ones, which means that the observed data provides information to the posteriors of both algorithms. Furthermore, we also observe that the PnP-SVGD posterior is narrower than the others, which we attribute to the constraint imposed by the DRUNet. In this case, DRUNet shapes the posterior to focus only on the high-probability region from which the high-resolution samples come. From the Bayesian perspective, the information provided by the prior and the constraint strongly influence the support of the posterior compared to the likelihood, which gains its support only from the observed data~\cite{spantini2015laplace, izzatullah2021laplace}. We also see that the width of the histograms increases in areas with larger variability–the ground truth falls inside the nonzero point-wise posterior interval, which confirms the benefits of the prior.

\subsection{Field data example: Volve model}
\begin{figure*}[!htb]
  \centering
  \includegraphics[width=\textwidth]{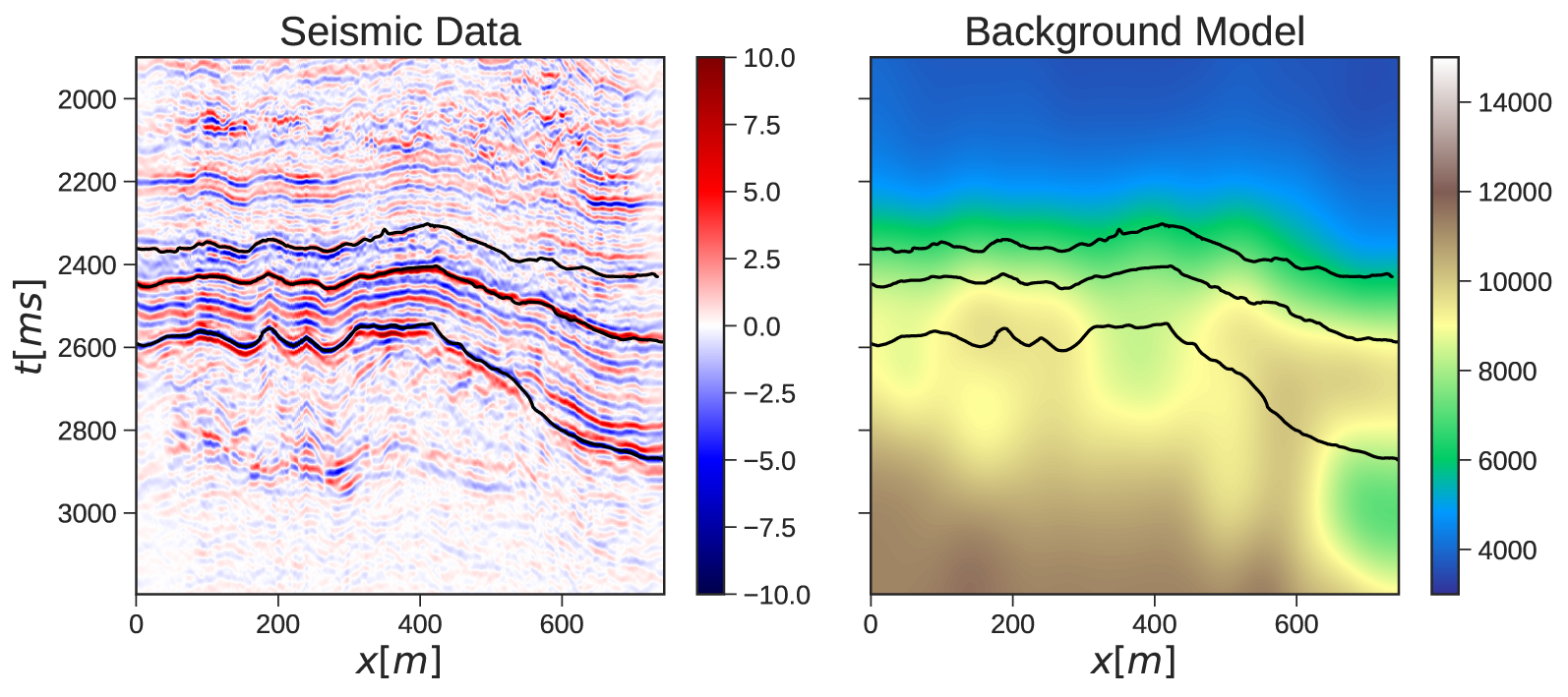}
  \caption{Volve model observed seismic data and acoustic impedance (AI) background model. The black lines represent the interpreted seismic horizons.}
  \label{fig:3-2-Volve-data}
\end{figure*}

In this second example, we demonstrate the potential of PnP-SVGD in performing posterior inference for a seismic field dataset. The dataset used in this numerical example is from the Volve oilfield, an oilfield in the central part of the North Sea, which was in production from 2008 to 2016~\cite{equinor2018data}. We refer the reader to~\cite{ravasi2022} for a detailed description of the Volve dataset and the data preparation process. The seismic data and AI background model are illustrated in Figure~\ref{fig:3-2-Volve-data}.

\begin{figure*}[!htb]
  \centering
  \includegraphics[width=\textwidth]{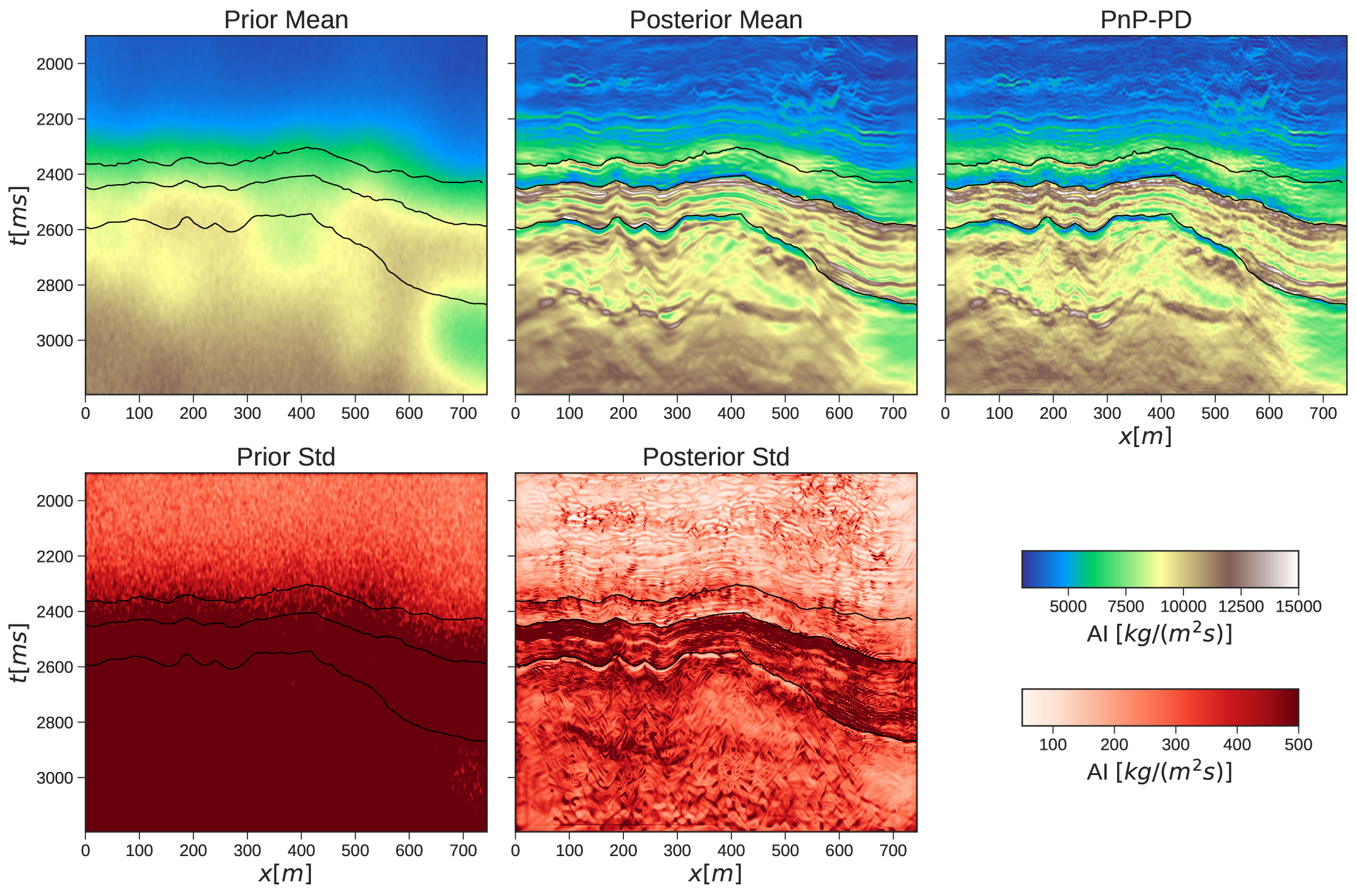}
  \caption{Volve model statistics. \textbf{Top row:} Mean models of the samples drawn from prior and posterior distributions, along with inverted model from the deterministic PnP-PD optimization procedure. \textbf{Bottom row:} The samples standard deviation of the samples drawn from prior and posterior distributions.}
  \label{fig:3-2-Volve-models}
\end{figure*}

We sample $100$ particles (samples) from an initial distribution, which we define to come from a Gaussian distribution with the AI background model in Figure~\ref{fig:3-2-Volve-data} as the mean and a diagonal covariance matrix $\mathbf{C}_{q_{0}} = 0.5 \mathbf{I}$. This choice is made to show that even with non-geologically plausible initial particles, PnP-SVGD can produce geologically realistic posterior samples. For the posterior inference procedure, we perform $50$ iterations of PnP-SVGD and collect all the $100$ updated particles as the posterior samples to perform statistical analysis at the end of inference process. We also apply the PnP-PD algorithm with $100$ iterations using the same denoiser (i.e., DRUNet) to compare with the posterior samples mean model. Statistics for the Volve experiment are shown in Figure~\ref{fig:3-2-Volve-models}.

\begin{figure*}[!htb]
  \centering
  \includegraphics[width=0.85\textwidth]{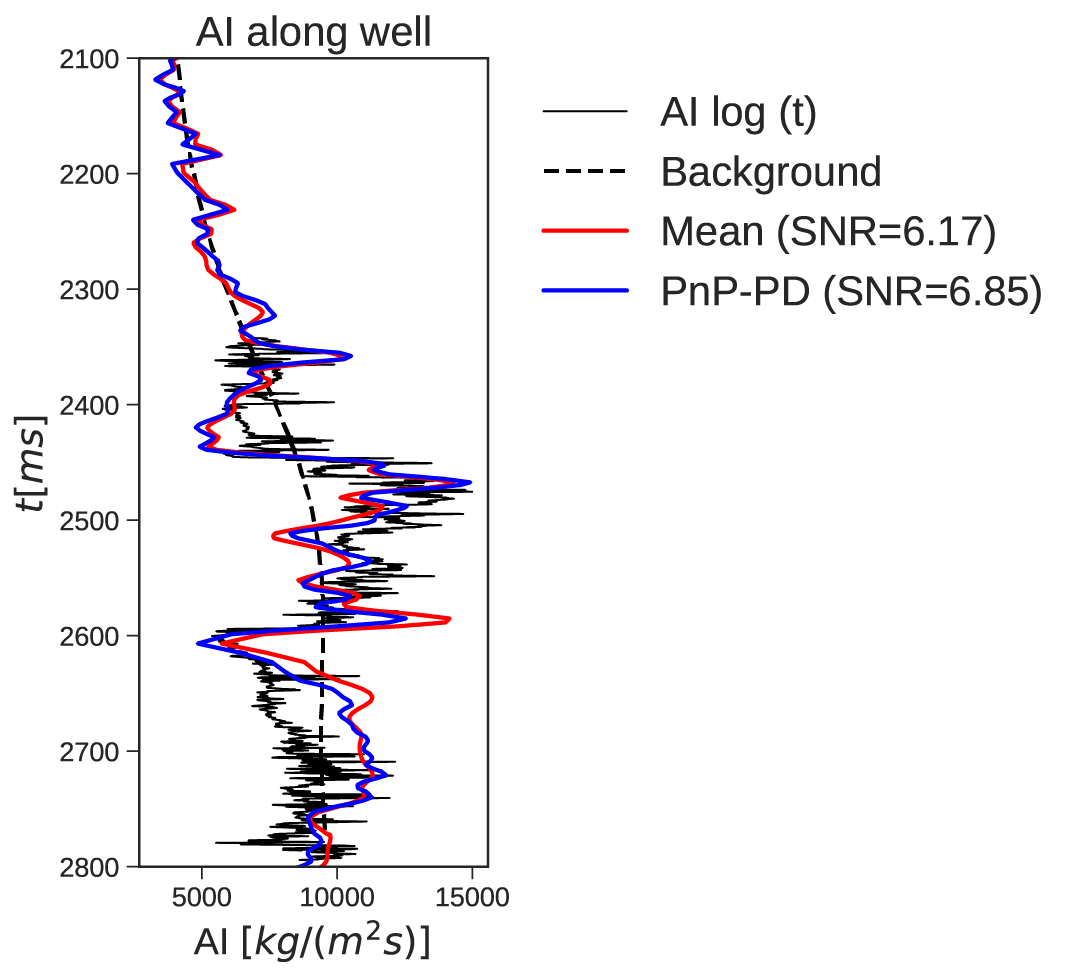}
  \caption{Volve model acoustic impedance comparison along well.}
  \label{fig:3-2-Volve-log}
\end{figure*}

\begin{figure*}[!htb]
  \centering
  \includegraphics[width=\textwidth]{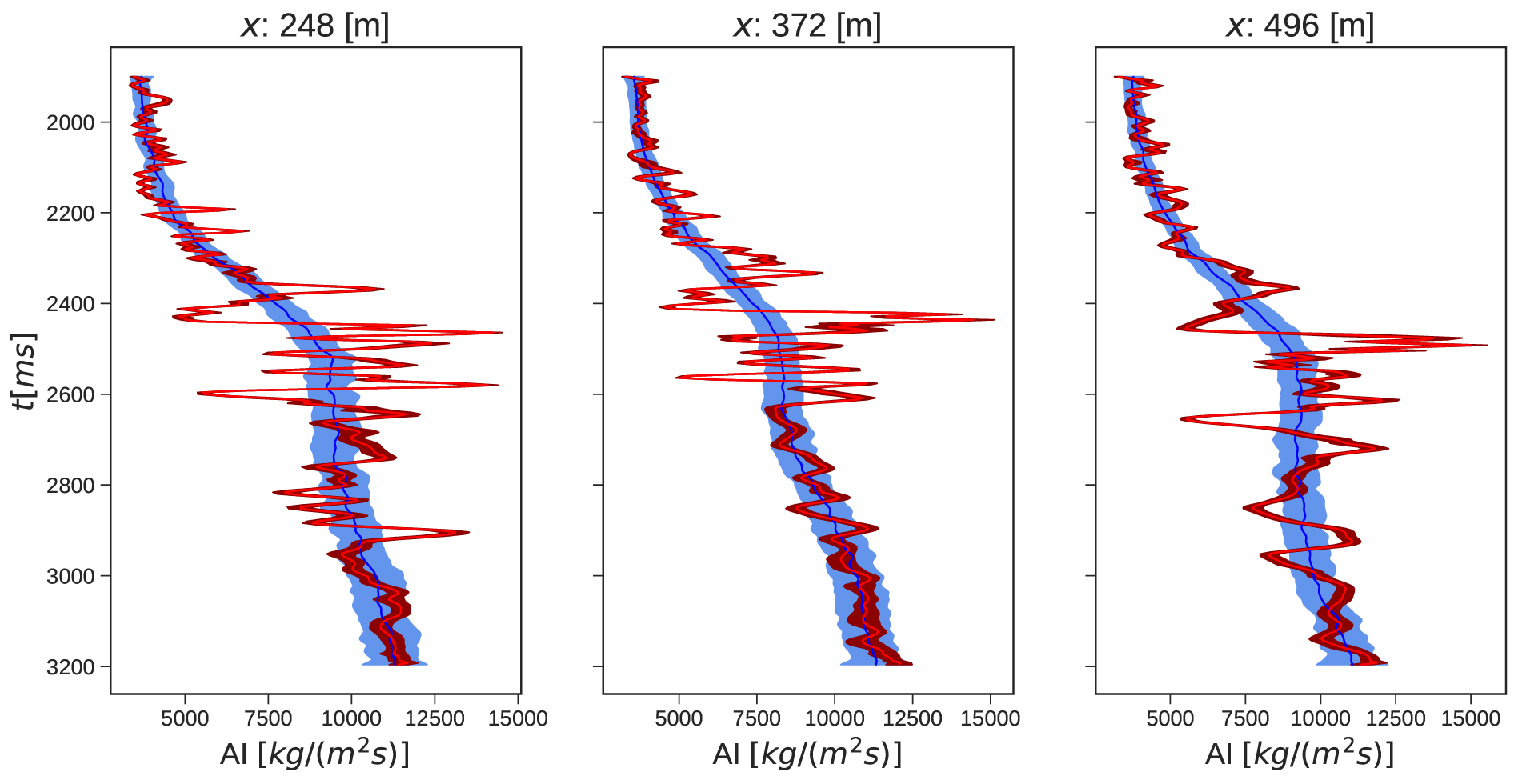}
  \caption{Volve model statistics per trace from PnP-SVGD procedure. The blue and red curves and regions represent the samples mean and 1-$\sigma$ credible-interval for the prior and posterior distributions, respectively.}
  \label{fig:3-2-Volve-traces}
\end{figure*}

Visually, the posterior sample mean obtained from the PnP-SVGD algorithm is comparable with the model estimate from the PnP-PD algorithm. However, in these simulations, we observe that the PnP-SVGD samples mean model provides a slightly lower SNR of $6.17$ dB compared to $6.85$ dB of PnP-PD along the well-log data in Figure~\ref{fig:3-2-Volve-log}.
According to Figure~\ref{fig:3-2-Volve-log}, PnP-SVGD overestimates the AI model between the range of $2550 - 2650$ ms compared to the PnP-PD. This result also tallies with the estimated posterior samples standard deviation in the bottom row of Figure~\ref{fig:3-2-Volve-models} and in Figure~\ref{fig:3-2-Volve-traces} for per trace statistics, where the high model uncertainty is recorded for AI starting from $2550$ ms.

\begin{figure*}[!htb]
  \centering
  \includegraphics[width=\textwidth]{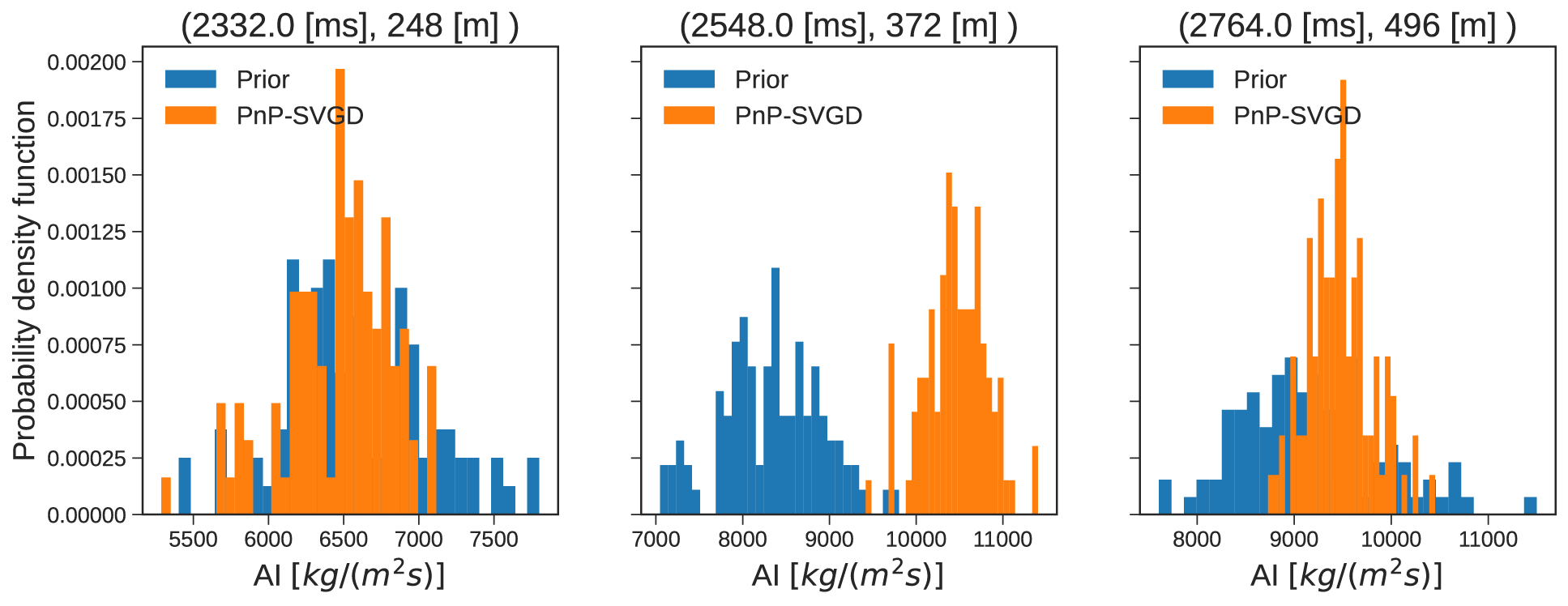}
  \caption{Volve model pointwise histogram for points located at \textbf{left:} ($2332$ [ms], $248$ [m]), \textbf{center:} ($2548$ [ms], $372$ [m]), and \textbf{right:} ($2764$ [ms], $496$ [m]).}
  \label{fig:3-2-Volve-hist}
\end{figure*}

\begin{figure*}[!htb]
  \centering
  \includegraphics[width=\textwidth]{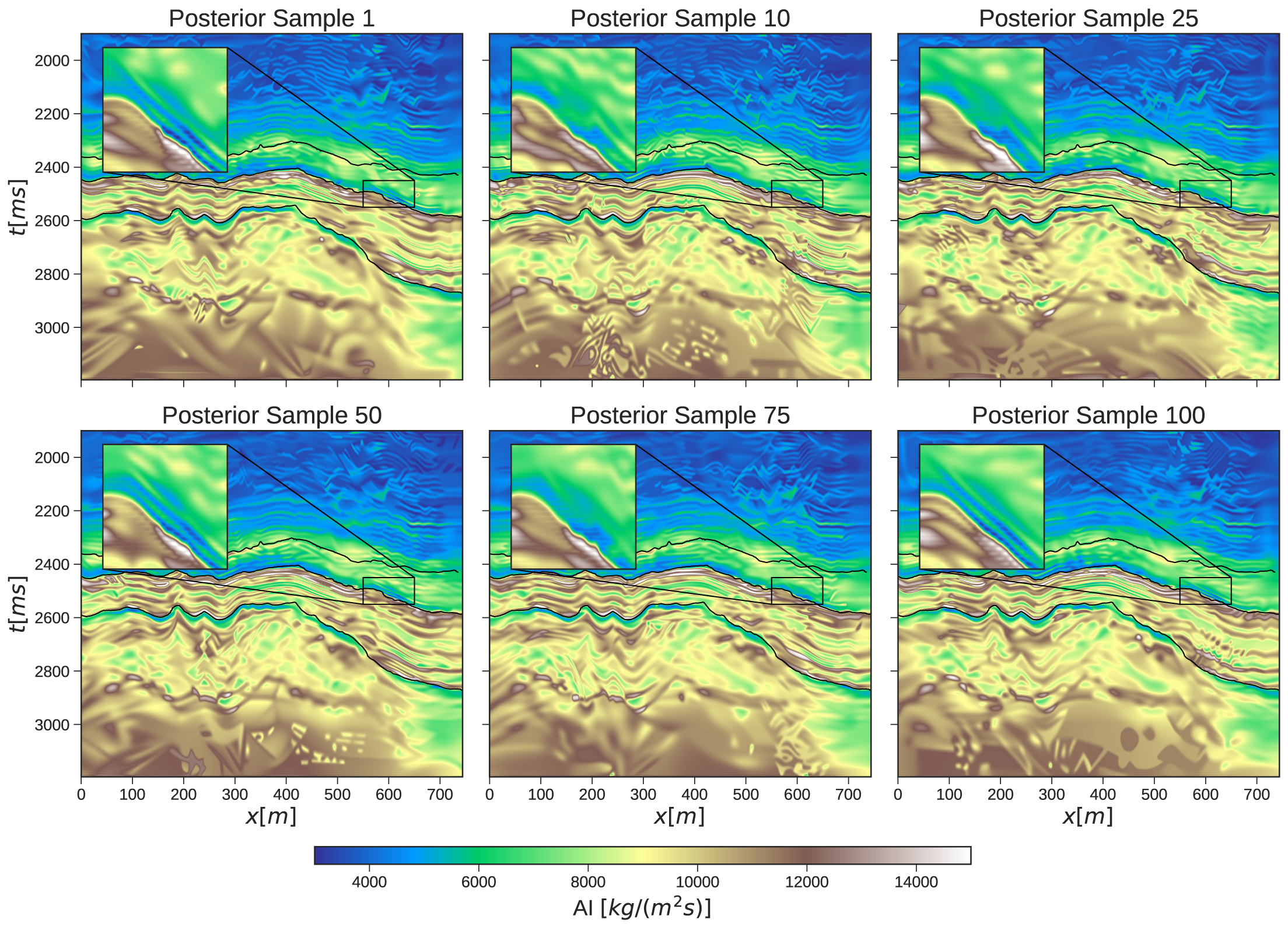}
  \caption{Volve model posterior samples from the PnP-SVGD procedure.}
  \label{fig:3-2-Volve-posterior-samples}
\end{figure*}

Furthermore, the region with the highest model uncertainty is observed within the middle and bottom lines of the seismic horizon. We observe high model variability in this region, as shown in Figure~\ref{fig:3-2-Volve-traces} for per-trace statistics and illustrated in Figure~\ref{fig:3-2-Volve-posterior-samples} for posterior sample variability. From the imaged seismic data, this region contains higher signal amplitude than other regions. Conventionally, we expect this region to provide high variance in the associated estimated AI model. However, similar to the arguments we made in the previous example, the inflation of variance in this region is attributed to the influence of the prior information and the constraint imposed on the posterior distribution. The regions with low signal amplitude are being compensated by the information from the prior and the constraint (alongside the observed data), resulting in low variance records. Nevertheless, this does not reflect the accuracy of the inverted model in those regions because only the observed data could validate such claims made by the prior and the constraint. To illustrate how the posterior is distributed, we also calculate point-wise histograms at three locations, as illustrated in Figure~\ref{fig:3-2-Volve-hist}.

\begin{landscape}
\begin{figure*}[!htb]
  \vspace{11\baselineskip}
  \includegraphics[width=1.3\textwidth, keepaspectratio]{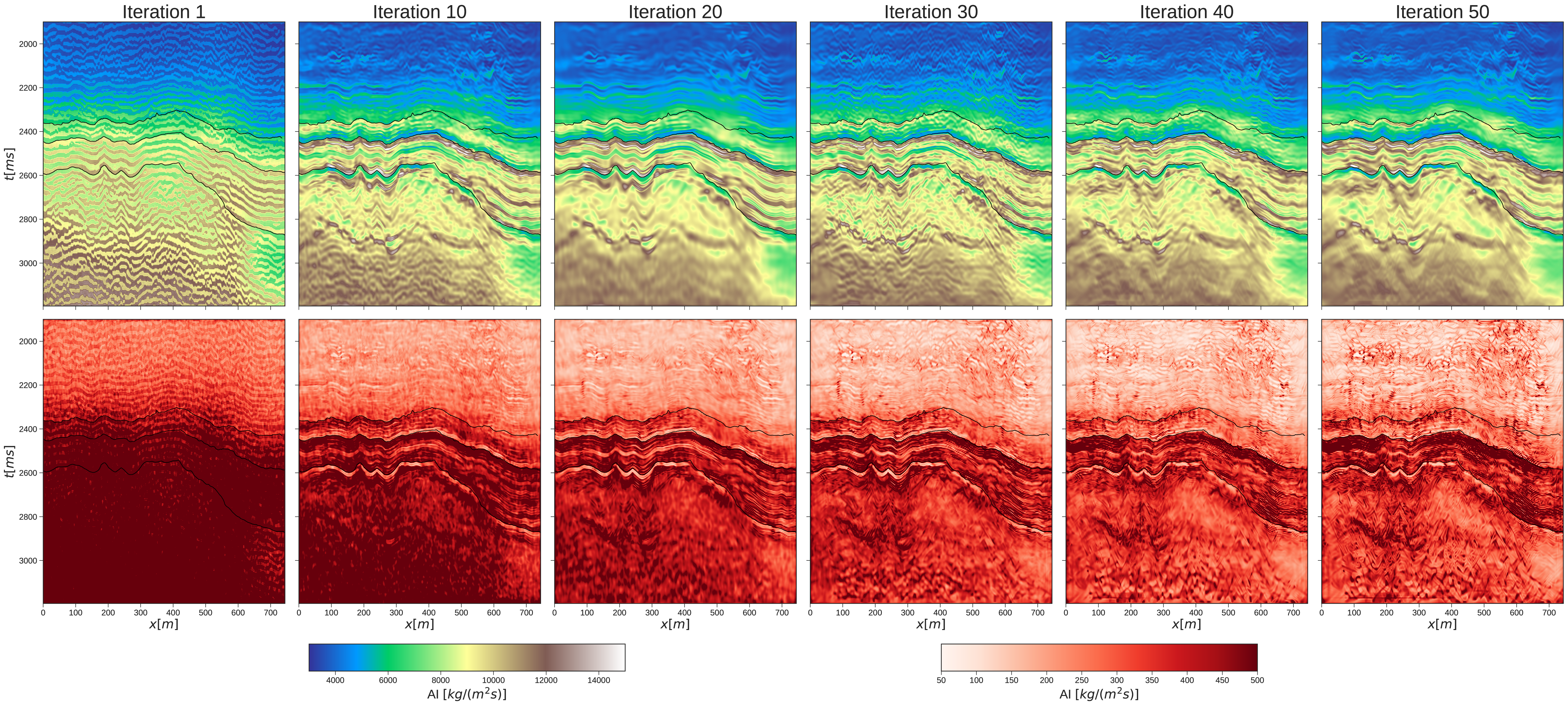}
  \caption{Volve model samples mean and standard deviation evolution per iteration for the PnP-SVGD procedure.}
  \label{fig:3-2-Volve-evo}
\end{figure*}
\end{landscape}

In Figure~\ref{fig:3-2-Volve-evo}, we show the evolution of samples mean and standard deviation evolution per iteration for the PnP-SVGD procedure. Qualitatively, the posterior samples from PnP-SVGD coupled with the DRUNet denoiser demonstrate the capability of the proposed algorithm to provide high-resolution samples from the posterior distribution as shown in Figure~\ref{fig:3-2-Volve-posterior-samples} and being robust towards field noise present in the observed data, where the field noise distribution is generally unknown.

\section{Discussion}
PnP-SVGD demonstrated its potential and capability to produce high-resolution and geologically plausible samples by sampling the regularized posterior distribution. In this section, we discuss the limitations of PnP-SVGD from three perspectives--computational, algorithm design, and hyperparameter tuning.

\subsection*{Computational aspect}
PnP-SVGD requires one forward and one adjoint operator evaluation per particle at each iteration, which grows the computational costs linearly with the number of particles. Furthermore, to obtain a good accuracy in posterior sampling, PnP-SVGD--SVGD in general--requires at least more particles than the number of model parameters, which would limit the scalability of PnP-SVGD on large-scale nonlinear inverse problems such as full waveform inversion (FWI)~\cite{zhang2020svgdseismic, zhang2022svgd}. In that case, we suggest the practitioners implement the stochastic gradient approach to reduce the computational burden at each iteration~\cite{huggins_random_2018, gorham_stochastic_2020}. Another alternative is to use the hybrid algorithm, such as stochastic SVGD (a hybrid between stochastic gradient MCMC and SVGD), which leverages the concept of multi-chain MCMC for posterior inference and substantially reduce the usage of a large number of particles as in SVGD-based algorithms~\cite[see, e.g.,][]{gallego_stochastic_2020, leviyev_stochastic_2022, zhang_3d_2022}.

Apart from this, the choice of deep denoiser also plays a crucial role in the efficiency of PnP-SVGD. The deep denoiser should not only provide a better regularization effect on the posterior samples, but it also should be computationally efficient. In such a case, we argue that a deep denoiser based on the current deep generative diffusion models~\cite[see, e.g.,][]{ho_denoising_2020, voleti_score-based_2022, yang_diffusion_2022} would not fit in this framework due to its high computational demands and slowness at the inference stage, despite its successes in producing high-quality images. 

\subsection*{Algorithm design}
The current development of PnP-SVGD is based on FBS, a family of the proximal algorithm in deterministic convex optimization. We favor this formulation for PnP-SVGD because of its close connection with the proximal LMC algorithm--based on the FBS formulation--which shares a common theoretical background~\cite{pereyra2016lmc, salim2019lmc, liu2021svgd}. Although PnP-SVGD can be extended by using other proximal algorithms, such as ADMM, the main concern lies in the theoretical aspect, where a direct adaptation from a deterministic optimization algorithm to the probabilistic inference setting is nontrivial. Asymptotically, the extended algorithm does not provide the samples from the target posterior, thus introducing bias in the statistical analysis. Several recent works demonstrated the success of practically adapting SVGD with the primal-dual algorithm~\cite[e.g.,][]{liu2021svgd, zhang2022svgd}. However, the theoretical support for such an algorithm remains an open research question.

\subsection*{Hyperparameter tuning}
In our PnP-SVGD formulation, there are several hyperparameters that influence its performance and efficiency, such as the choice of kernel, step size, and noise parameter of the deep denoiser, $\sigma$. For the choice of kernel, Gorham and Mackey~\cite{gorham2017stein} argued that SVGD, PnP-SVGD included, with RBF kernel, may not be efficient in high dimensions and suggested using inverse multiquadric (IMQ) kernel, which does not suffer from this issue. However, this suggestion assumes a fixed bandwidth, which does not hold for the RBF kernel equipped with the median trick, which can better adapt to the samples at each iteration. We leave the study of the proper choice of kernels to future works. As for the step size, it controls the convergence of PnP-SVGD and SVGD, where a proper step size decaying strategy would help the algorithms converge faster. We use cosine annealing~\cite{loshchilov_sgdr_2017} as the step size decaying strategy in the demonstrated examples. A poor choice of step size would increase the computational costs as the number of iterations would also increase, thus increasing the number of forward and adjoint operator evaluations.

On the other hand, the noise parameter of the deep denoiser influences the resolution of the posterior samples. A large value of $\sigma$ would produce over-smoothed samples, which are not geologically plausible. Since the deep denoiser acts on the samples at each iteration, $\sigma$ should be chosen as small enough to provide a good compounding regularization effect on the posterior samples at each iteration.

\section{Conclusions}
We presented a regularized variational inference framework. This computational framework performs posterior inference by regularizing the KL divergence loss with an implicit regularization implemented via a CNN-based denoiser. We propose the PnP-SVGD algorithm as a way to sample from a regularized target posterior. Our numerical results on the post-stack seismic inversion suggest that PnP-SVGD coupled with the DRUNet denoiser has the capability of producing high-resolution samples representative of the subsurface structures for inference purposes. PnP-SVGD also outperformed vanilla SVGD in terms of the posterior samples quality and robustness in the inference procedure with respect to noise present in the data. With such capabilities, the samples from PnP-SVGD can be used further for post-inference tasks (e.g., reservoir modelling and history matching) because of their high-resolution and realistic representation of the subsurface structures.
\section*{Acknowledgment}
This publication is based on work supported by the King Abdullah University of Science and Technology (KAUST). The authors thank the DeepWave sponsors for supporting this research and also thank \href{https://www.equinor.com/energy/volve-data-sharing}{Equinor and the Volve license partners} for providing access to the data used in one of the examples. 
\section*{Code Availability}
The data and accompanying codes that support the findings of this study will be openly available at \href{https://github.com/DeepWave-KAUST/PnPSVGD}{https://github.com/DeepWave-KAUST/PnPSVGD}. 

\printbibliography

\end{document}